\documentclass[runningheads]{llncs}

\usepackage{eccv}

\usepackage{eccvabbrv}

\usepackage{booktabs}
\usepackage{multirow}
\usepackage{graphicx}
\usepackage{xcolor,colortbl}
\usepackage{makecell}
\usepackage{wrapfig}
\usepackage{caption}
\usepackage[ruled,vlined,linesnumbered]{algorithm2e}

\usepackage{ulem}
\usepackage{longtable}
\usepackage{lineno}
\definecolor{bestred}{RGB}{214, 39, 40}
\definecolor{secondblue}{RGB}{31, 119, 180}
\definecolor{rowgray}{gray}{0.95} 

\newcommand{\best}[1]{\textbf{\textcolor{bestred}{#1}}}
\newcommand{\second}[1]{\underline{\textcolor{secondblue}{#1}}}

\usepackage[accsupp]{axessibility}  

\usepackage{hyperref}

\usepackage{orcidlink}
\usepackage{tcolorbox}

\makeatletter
\newcommand{\printfnsymbol}[1]{%
  \textsuperscript{\@fnsymbol{#1}}%
}
\makeatother

\begin{document}

\title{Decoder-Free Distillation for Quantized Image Restoration} 

\titlerunning{QDR}

\author{S. M. A. Sharif\inst{1}\thanks{Equal contribution} \and
Abdur Rehman\inst{1}\printfnsymbol{1} \and
Seongwan Kim\inst{1} \and
Jaeho Lee\inst{1}}

\authorrunning{S. M. A. Sharif et al.}

\institute{Opt-AI Inc., Seoul, South Korea\\
\email{\{sharif, abdur, swan.kim, jaeho.lee\}@opt-ai.kr}}

\maketitle

\vspace{-.6cm}

\begin{figure}[!htb] 
  \centering
  
  \begin{minipage}[t]{0.35\linewidth}
    \vspace{0pt} 
    \includegraphics[width=\linewidth, height=3.8cm]{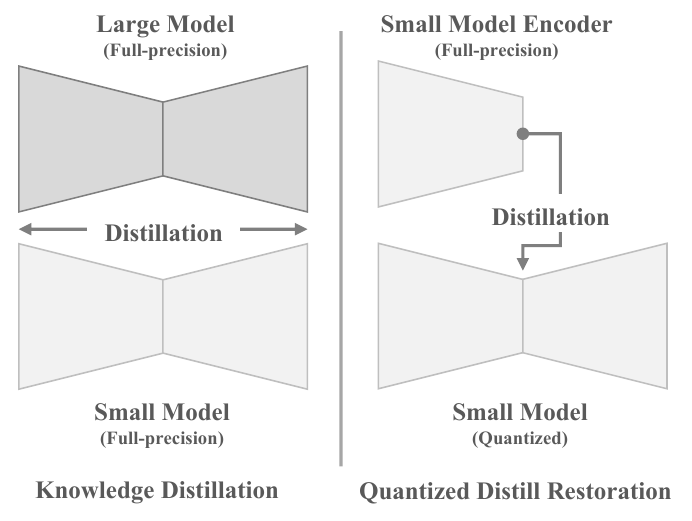}
  \end{minipage}
  %
  \begin{minipage}[t]{0.38\linewidth}
    \vspace{0pt}
    \includegraphics[width=\linewidth, height=3.8cm]{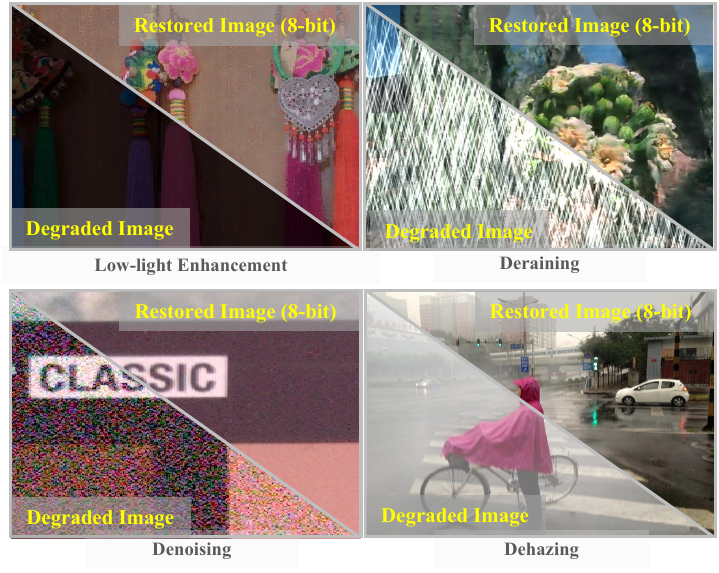}
  \end{minipage}
  %
  \begin{minipage}[t]{0.25\linewidth}
    \vspace{-1pt}
    \includegraphics[width=\linewidth, height=2.5cm]{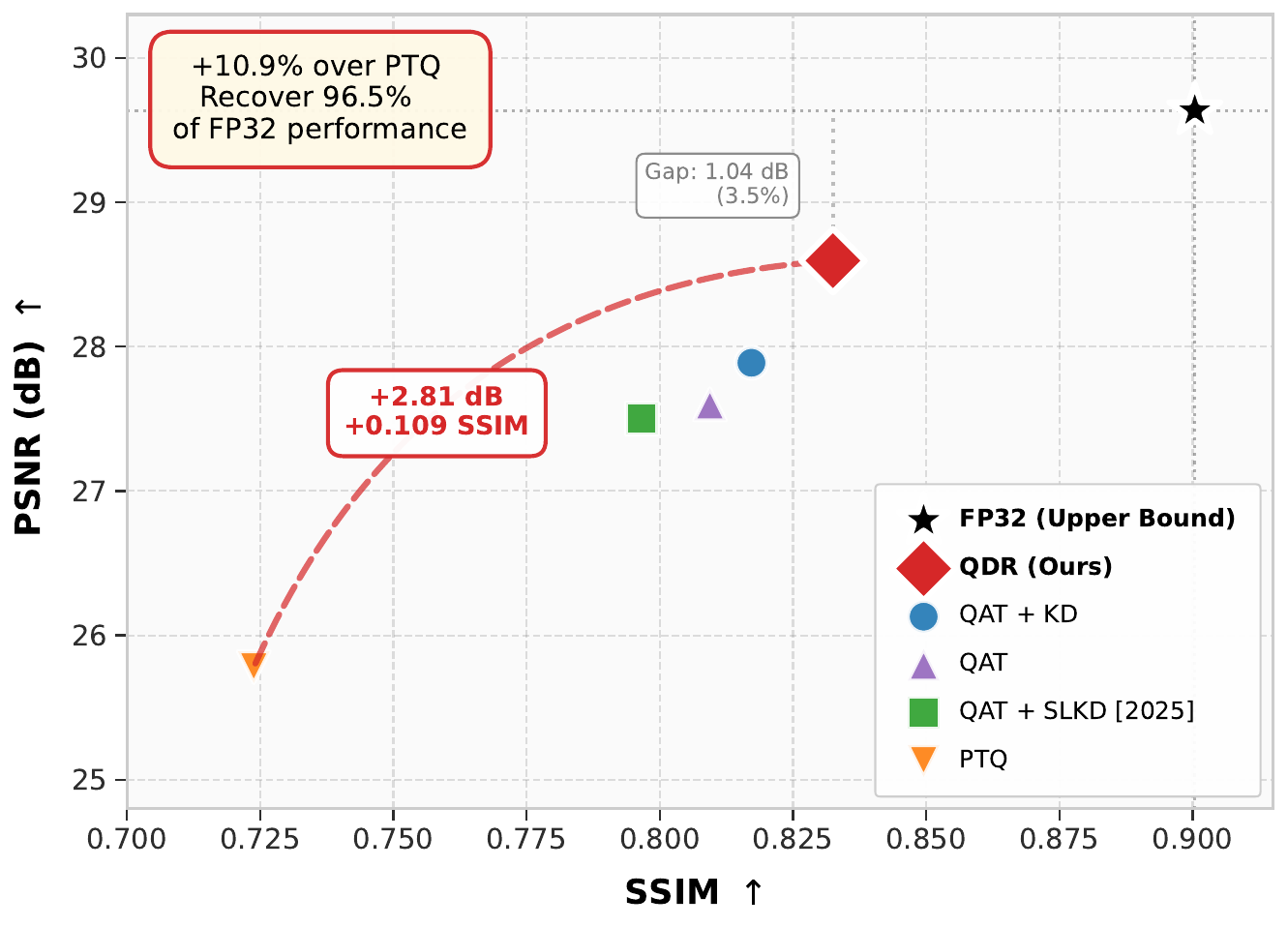}\\[0.01cm]
    \includegraphics[width=\linewidth, height=1.2cm]{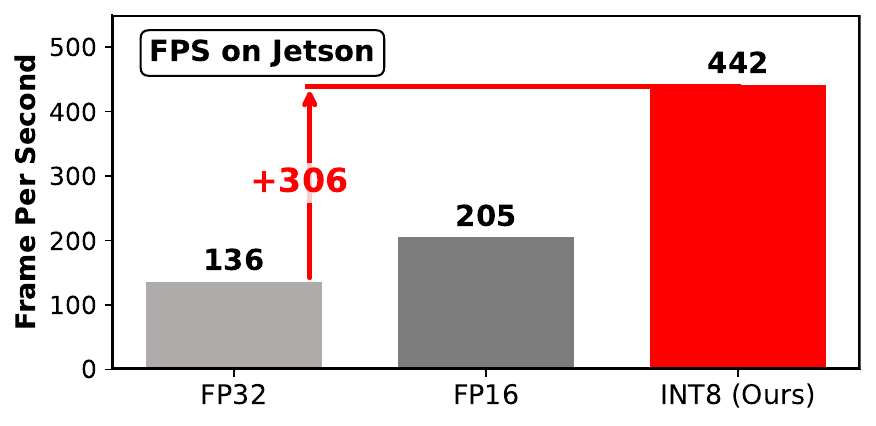}
  \end{minipage}
  
  \vspace{-0.2cm} 
  \caption{Comparison of standard Distillation and our proposed Quantization-aware Distill Restoration (QDR) paradigm (\textbf{Left}). QDR enables high-quality restoration across diverse degradations (low-light, deraining, denoising, dehazing) (\textbf{Middle}). Our method recovers $\sim$96.5\% of FP32 and significantly faster FPS (442) on the edge board (\textbf{Right}).}
  \label{fig:intro}
  
\end{figure}

\vspace{-1.1cm}
\begin{abstract}
Quantization-Aware Training (QAT), combined with Knowledge Distillation (KD), holds immense promise for compressing models for edge deployment. However, joint optimization for precision-sensitive image restoration (IR) to recover visual quality from degraded images remains largely underexplored.  Directly adapting QAT-KD to low-level vision reveals three critical bottlenecks: teacher-student capacity mismatch, spatial error amplification during decoder distillation, and an optimization "tug-of-war" between reconstruction and distillation losses caused by quantization noise. To tackle these, we introduce Quantization-aware Distilled Restoration (QDR), a framework for edge-deployed IR. QDR eliminates capacity mismatch via FP32 self-distillation and prevents error amplification through Decoder-Free Distillation (DFD), which corrects quantization errors strictly at the network bottleneck. To stabilize the optimization tug-of-war, we propose a Learnable Magnitude Reweighting (LMR) that dynamically balances competing gradients. Finally, we design an Edge-Friendly Model (EFM) featuring a lightweight Learnable Degradation Gating (LDG) to dynamically modulate spatial degradation localization. Extensive experiments across four IR tasks demonstrate that our Int8 model recovers ~96.5\% of FP32 performance, achieves 442 frames per second (FPS) on an NVIDIA Jetson Orin, and boosts downstream object detection by 16.3 mAP. 
\end{abstract}

\section{Introduction}
\label{sec:intro}

Downstream tasks such as object detection and segmentation in real-world edge vision applications (EVA) are often hindered by environmental factors, including low light, haze, and rain. IR methods aim to recover perceptual quality and improve performance on these core tasks\cite{li2018benchmarking, yang2017deep, abdelhamed2018high, wei2018deep}. However, state-of-the-art (SOTA) IR models\cite{cai2023retinexformer, sharif2026illuminating, zamir2022restormer, wang2022uformer} are computationally intensive and incur high memory overhead, making them exceedingly difficult to deploy on resource-constrained edge platforms, including smartphones, drones, IoT sensors, and autonomous systems. Although model compression techniques such as quantization \cite{wang2021fully, li2020pams, hong2022daq, zhong2022dynamic, ayazoglu2021extremely} and KD \cite{zhang2026dlienet, zhang2025soft, zhang2025knowledge, zhou2025dynamic} offer a path forward, achieving an optimal trade-off between efficiency and performance remains an open challenge.

To enable real-time inference on edge hardware (e.g., NPUs and DSPs), integer quantization is highly favored for its ability to accelerate computation and reduce memory bandwidth. Unfortunately, IR is a dense regression task that is notoriously sensitive to numerical precision. Standard Post-Training Quantization (PTQ) and even QAT introduce severe quantization noise, leading to restored images with noticeable artifacts. In high-level vision tasks (e.g., classification, detection), these quantization errors are often tackled by combining QAT with KD \cite{zhu2023quantized, pham2023cmtkd, kim2019qkd}, using a full-precision teacher to transfer robust knowledge to a quantized student. However, directly adapting this paradigm to low-level vision is non-trivial; while high-level tasks primarily distill semantic abstractions, IR requires the strict preservation of fine-grained spatial details, which are easily corrupted by quantization. Consequently, despite its immense promise for edge deployment, the joint optimization of QAT and KD, which we term \textbf{QDR}, remains largely unexplored for low-level IR tasks.

Although QDR shows immense potential, it faces three critical, underexplored challenges. First, teacher selection: our analysis reveals that in low-level vision, transferring knowledge from a large, heterogeneous teacher to a heavily quantized student often fails due to severe capacity mismatch; the student simply cannot mimic the teacher's complex feature space \cite{yang2022masked}. Second, distillation localization: in standard encoder-decoder IR architectures, KD is typically applied across all stages or explicitly at the decoder. However, we observe that under quantization noise, forcing decoder distillation is fundamentally flawed; it compels the network to reconstruct clean outputs from corrupted bottleneck features, amplifying quantization errors during upsampling. Third, loss balancing: we find that jointly optimizing reconstruction and distillation losses creates a tug-of-war \cite{rehman2025punching}. Quantization introduces parameter-dependent, heteroskedastic gradient perturbations, making standard joint optimization highly unreliable \cite{zhao2024self}. \textit{In a nutshell, to successfully exploit QDR, three critical questions need to be answered: Which teacher provides the most robust knowledge? Where should distillation occur within the network? And how can we stabilize the joint optimization?}


To address these questions, we introduce a novel QDR framework that directly resolves these bottlenecks. First, to eliminate capacity mismatch, we leverage self-distillation, with the full-precision (FP32) network serving as the teacher. Second, to prevent error amplification during QDR, we introduce DFD, as shown in Fig. \ref{fig:intro}(Left). Rather than forcing distillation at the decoder, DFD applies supervision strictly at the network bottleneck, correcting the feature distribution at its source. Notably, we demonstrate that aligning the bottleneck naturally improves downstream layer alignment, enabling the quantized decoder to achieve near-perfect spatial alignment with the FP32 target without requiring any complex distillation mechanisms. Finally, to address instability in joint optimization, we introduce an LMR. Our LMR leverages exponentially smoothed gradient magnitudes to dynamically modulate the reciprocal weighting between the reconstruction and distillation objectives via learnable parameters, thereby mitigating the oscillatory effects of quantization noise.

Furthermore, to maximize hardware efficiency and QDR compatibility, we design an EFM comprising quantization-friendly operators. Rather than relying on heavy, quantization-sensitive attention mechanisms, EFM achieves high-quality restoration with a lightweight LDG. Our LDG dynamically modulates spatial features based on localized degradations, adding minimal computational overhead.

In summary, our main contributions are as follows:

\begin{itemize}

    \item We identify the pitfalls of standard distillation in quantized image restoration and propose DFD, demonstrating that explicit bottleneck alignment naturally heals decoder representations under quantization noise.

    \item We introduce a LMR that robustly balances reconstruction and distillation gradients, stabilizing joint optimization by accounting for quantization-induced gradient perturbations.

    \item We design a highly efficient EFM  with LDG to handle spatially varying corruptions with minimal computational overhead.

    \item Extensive experiments across four restoration tasks (denoising, low-light enhancement, deraining, and dehazing) demonstrate that our method achieves SOTA Int8 performance. Our model effectively recovers $\sim$96.5\% of FP32 performance, runs at 442 FPS on an NVIDIA Jetson Orin, and significantly improves downstream object detection (+ 16.3 mAP on ExDark \cite{loh2019getting}) in degraded environments (see Fig. \ref{fig:intro}).

\end{itemize}

\section{Related Work}
\subsection{Quantization for Image Restoration}

Quantization is a fundamental step for model compression, particularly for edge deployment. Early IR QAT works \cite{wang2021fully, li2020pams, hong2022daq, zhong2022dynamic, ayazoglu2021extremely} focused on designing edge-friendly models. Recently, quantization has expanded to more complex architectures, including PTQ for Diffusion models \cite{wu2024ptq4dit} and State Space Models \cite{pierro2024mamba}, alongside advancements like PREFILT \cite{makhov2025prefilt}, which mitigates internal noise in fully quantized networks via input prefiltering. Despite these advances, applying hardware-friendly integer compression to diverse environmental degradations (e.g., rain, haze) remains underexplored. Notably, approaches like VQCNIR \cite{zou2024vqcnir} employ vector quantization for night-time priors; they focus on discrete feature representation rather than the integer weight and activation compression required for edge NPUs. Notably, existing studies largely rely on simulated quantization and overlook this joint optimization QDR.

\subsection{Knowledge Distillation for Image Restoration}

Adapting KD for dense IR requires preserving fine-grained spatial details \cite{zheng2021learning}. Recent specialized frameworks tackle complex IR tasks via attention mechanisms (DLIENet \cite{zhang2026dlienet}, SKD \cite{zhang2025soft}), contrastive learning (DCKD \cite{zhou2025dynamic}), or dual-teacher architectures (SLKD \cite{zhang2025knowledge}) that explicitly force the student decoder to learn image reconstruction. However, directly applying these standard paradigms to \textit{quantized} IR models introduces severe bottlenecks. Existing methods assume a full-precision student; they force multi-stage or decoder-level distillation. Under integer quantization, this compels the network to reconstruct clean outputs from heavily corrupted bottleneck features, severely amplifying quantization errors during upsampling. To overcome this, our proposed DFD fundamentally diverges by applying supervision strictly at the bottleneck, naturally healing representations without the error amplification inherent in traditional KD.

\subsection{Quantization with Knowledge Distillation}
Recent research extensively leverages KD to mitigate accuracy drops in low-precision models. Existing efforts primarily targeted high-level discriminative vision tasks, introducing phased coordination (QKD \cite{kim2019qkd}), stochastic bit precision (SPEQ \cite{boo2021stochastic}), and quantized or multi-teacher frameworks (QFD \cite{zhu2023quantized}, CMT \cite{pham2023cmtkd}). Beyond discriminative tasks, QAT-KD has recently expanded to generative models, such as Q-VDiT \cite{feng2025q} for video diffusion transformers. However, a fundamental challenge in these existing QAT-KD frameworks is the optimization "tug-of-war" between the task-specific reconstruction loss and the distillation loss.  SQAKD \cite{zhao24d} attempts to balance this in classification by ignoring task-specific loss. GoR counters it with learnable balancing and generalizes it on numerous high-level tasks. However, IR is highly precision-sensitive, forcing a quantized student to simultaneously balance image reconstruction with teacher feature matching, which leads to severe spatial error amplification. To overcome this, our work pioneers QAT-KD for IR, resolving the traditional tug-of-war by fundamentally restructuring the distillation paradigm.

\section{Methods}

\subsection{Problem Formulation}

Let $\mathbf{x} \in \mathcal{X}$ denote a degraded image and $\mathbf{y} \in \mathcal{Y}$ its clean target from dataset $\mathcal{D}=\{(\mathbf{x}_i,\mathbf{y}_i)\}_{i=1}^{N}$. A restoration network $f_{\theta}$ learns $\hat{\mathbf{y}} = f_{\theta}(\mathbf{x})$ by minimizing a reconstruction objective $\mathcal{L}_{\text{rec}}=\ell(\hat{\mathbf{y}},\mathbf{y})$, where $\ell(\cdot,\cdot)$ may combine $\ell_1$/$\ell_2$, perceptual, or frequency-domain losses. For edge deployment, we adopt QAT, which simulates low-bit arithmetic during training via a quantizer $\mathcal{Q}(\cdot)$ applied to weights and/or activations: $\hat{\mathbf{y}}_{S}=f_{\mathcal{Q}(\theta_S)}(\mathbf{x}), 
\mathcal{L}_{\text{QR}}=\ell\!\left(\hat{\mathbf{y}}_{S},\mathbf{y}\right)$. KD \cite{hinton2015distilling} transfers knowledge from a full-precision teacher $f_{\theta_T}$ to a lightweight student $f_{\theta_S}$. Given teacher prediction $\hat{\mathbf{y}}_{T}=f_{\theta_T}(\mathbf{x})$, the distillation objective is: $
\hat{\mathbf{y}}_{T}=f_{\theta_T}(\mathbf{x}), \mathcal{L}_{\text{KD}} = d\!\left(\hat{\mathbf{y}}_{S},\hat{\mathbf{y}}_{T}\right)$ where $d(\cdot,\cdot)$ may operate on outputs and/or intermediate representations. Since QAT and KD are complementary, we jointly optimize them under a unified \emph{quantization-aware distilled restoration} (QDR) objective:



\begin{equation}
\mathcal{L}_{\text{QDR}}=\mathcal{L}_{\text{QR}}+\lambda\,\mathcal{L}_{\text{KD}}
\end{equation}

We address three key challenges in QDR: \textbf{(i) Distillation localization:} distilling all encoder--decoder stages is often unnecessary and suboptimal under quantization noise; we systematically study \emph{where} to apply KD to maximize the quality--efficiency trade-off.\textbf{(ii) Teacher choice:} which type of teacher is most effective in QDR settings. \textbf{(iii) Loss balancing:} jointly optimizing $\mathcal{L}_{\text{QR}}$ and $\mathcal{L}_{\text{KD}}$ is unstable due to competing gradients, making $\lambda$ non-trivial to set. Critically, the gradient w.r.t.\ quantized student parameters decomposes as:

\begin{equation}
\label{eq:qat_grad}
\nabla_{\mathcal{Q}(\theta_S)}\mathcal{L}_{\text{QR}}
= \mathbf{g}(\theta_S) + \boldsymbol{\xi}(\theta_S)
\end{equation}

where $\mathbf{g}(\theta_S)$ is the clean full-precision gradient and $\boldsymbol{\xi}(\theta_S)$ is a parameter-dependent, biased, heteroskedastic quantization perturbation. Since the scale and distribution of $\boldsymbol{\xi}$ differ across the two losses, standard adaptive weighting schemes \cite{lu2021rw, li2021dynamic} can be miscalibrated or even counteract the two losses \cite{zhao2024self}. Building on \cite{rehman2025punching}, we replace the fixed scalar $\lambda$ with two learnable scalars $\lambda_{\text{rec}}$ and $\lambda_{\text{kd}}$ that adaptively reweight both terms:

\begin{equation}
\label{eq:gor_eq}
\mathcal{L}_{\text{QDR}}
= \frac{\lambda_{\text{rec}}}{\lambda_{\text{kd}}}\,\mathcal{L}_{\text{QR}}
+ \frac{\lambda_{\text{kd}}}{\lambda_{\text{rec}}}\,\mathcal{L}_{\text{KD}}
\end{equation}

We further extend this strategy for quantized image restoration in a later section, yielding more stable optimization and stronger restoration quality.

\subsection{ Quantization-aware Distilled Restoration (QDR)}
\begin{figure}
    \centering
    \includegraphics[width=.9\linewidth]{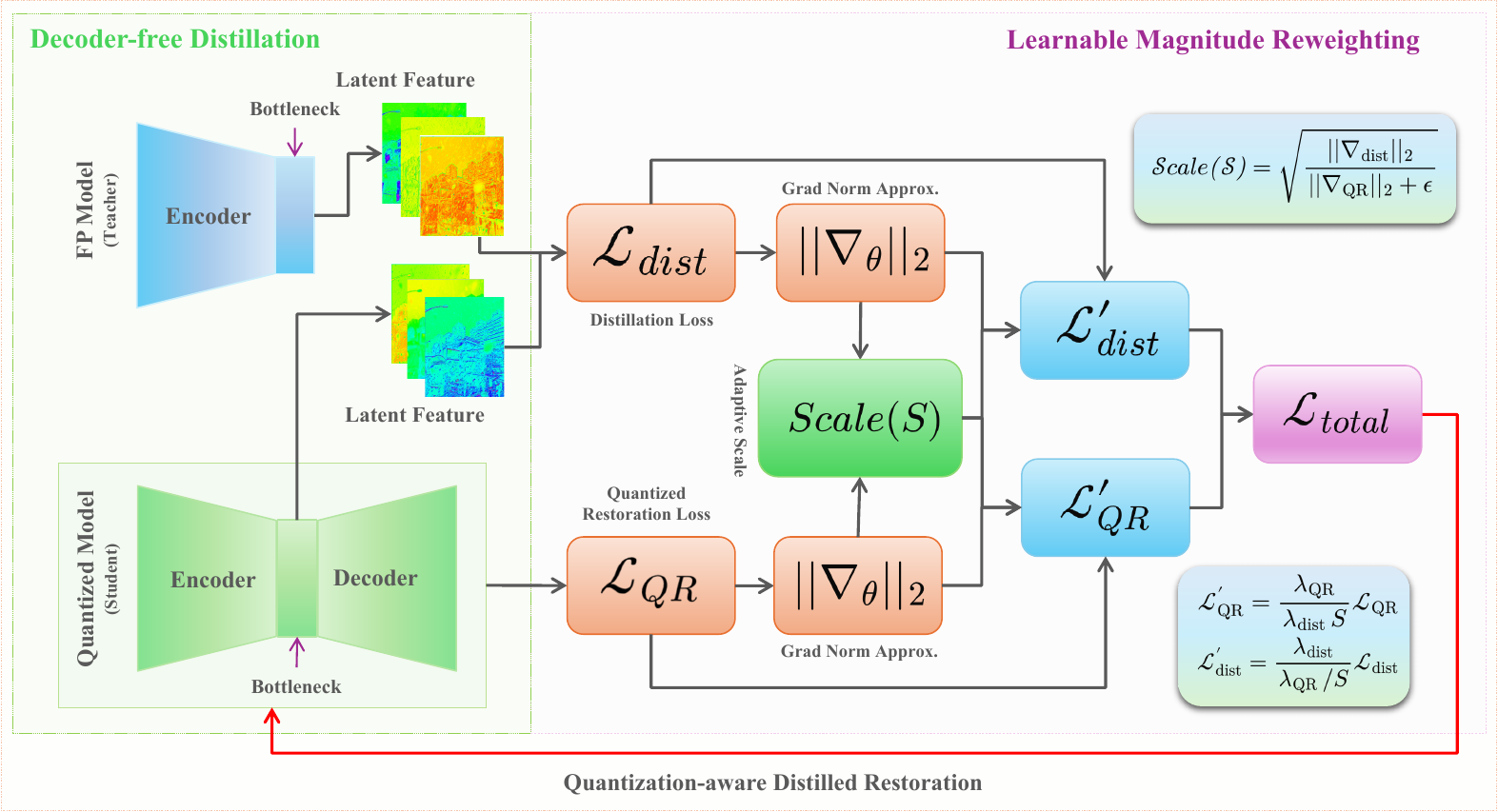}
    \caption{Overview of the proposed QDR framework, which consists of two components: DFD and LMR. The DFD architecture ensures effective knowledge distillation under quantization, while the LMR module dynamically reweights the magnitudes of reconstruction and distillation losses to improve training stability and performance.}
    \label{fig:overview}
    \vspace{-0.5cm}
\end{figure}

\begin{wrapfigure}{r}{0.6\linewidth}
\vspace{-.7cm}
  \centering
  \includegraphics[width=\linewidth]{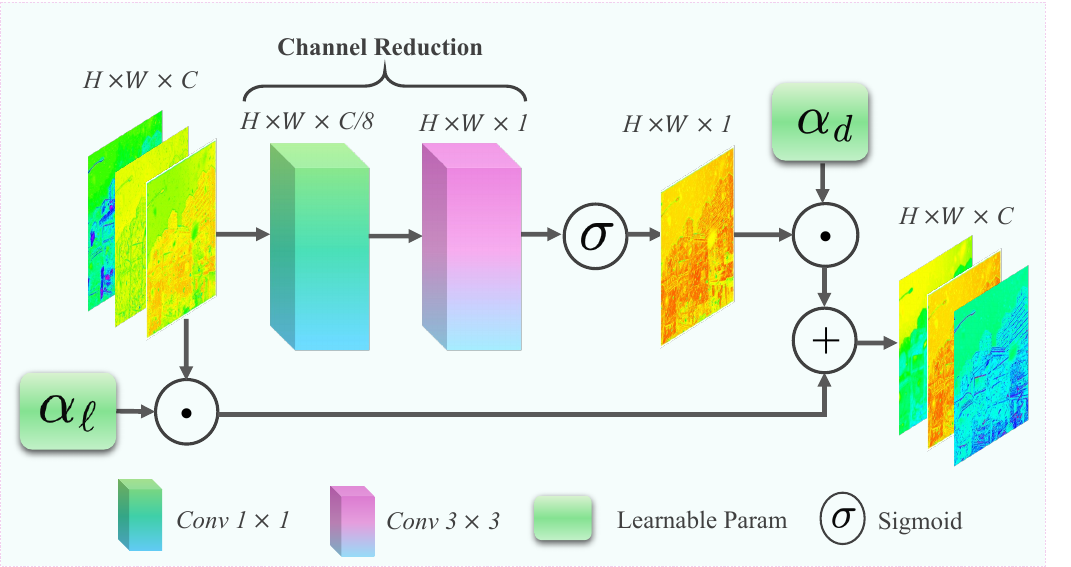}
    \caption{Architecture of our proposed Learnable Degradation Gating module containing two trainable scalers $[\alpha_{\ell}, \alpha_d]$ for controlling the fusion of Degradation importance map with residual features.}
    \label{fig:LDG}
  \vspace{-.8cm}
\end{wrapfigure}

We address the above stated three challenges in QDR by proposing DFD and LMR (Fig.\ref{fig:overview}). To fully exploit QDR, we also propose a lightweight EFM with LDG for effective image restoration (Fig. \ref{fig:LDG}) and evaluate QDR performance on real hardware.

\subsubsection{Decoder-free Distillation}
To address the first two bottlenecks in quantized IR, our DFD explicitly determines both the teacher choice and the optimal distillation location.

\paragraph{Distillation Localization:}

\begin{wrapfigure}{r}{0.54\linewidth} 
    \centering
    \vspace{-0.8cm}
    \begin{subfigure}{0.48\linewidth}
        \centering
        \includegraphics[width=3.2cm, height=3cm]{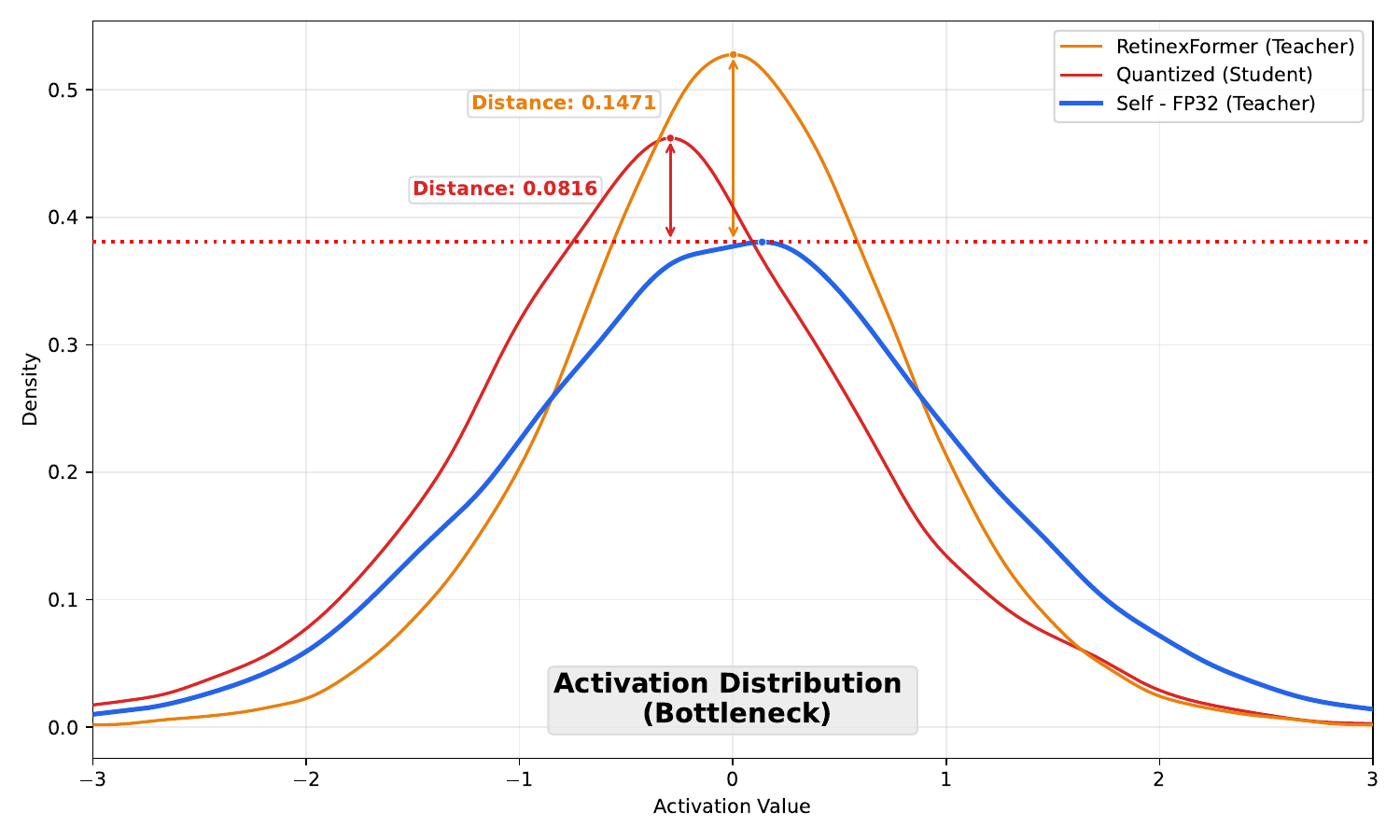}
        \caption{Bottleneck}
        \label{fig:bn_hist}
    \end{subfigure}
    \begin{subfigure}{0.48\linewidth}
        \centering
        \includegraphics[width=3.2cm, height=3.cm]{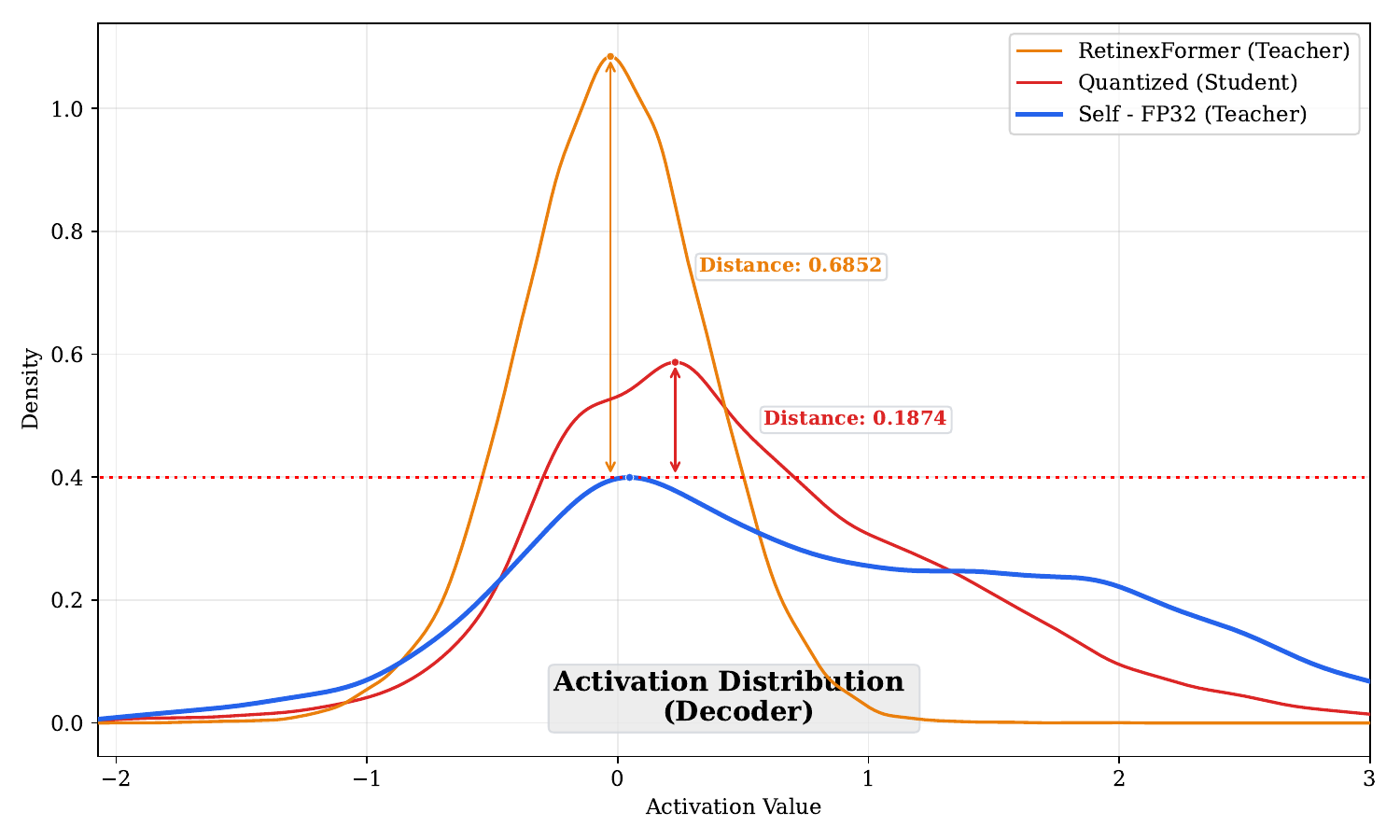}
        \caption{Decoder}
        \label{fig:dec_hist}
    \end{subfigure}
    
    \caption{Activation histograms at the bottleneck and decoder layers, illustrating the distribution differences between the FP32 and quantized models across different architectures.}
    \label{fig:activation_histograms}
    \vspace{-1.cm}
\end{wrapfigure}

We factorize the quantized student as $\hat{\mathbf{y}}_S = D_S(\mathbf{z}_S)$, where $\mathbf{z}_S = B_S(E_S(\mathbf{x}))$ is the bottleneck representation and $D_S$, $B_S$, $E_S$ denote the decoder, bottleneck, and encoder of $f_{\mathcal{Q}(\theta_S)}$. Acting as an \emph{information choke point}, $\mathbf{z}_S$ must retain task-relevant content while discarding redundant variation, making its distribution stable across training schemes unlike late decoder features, which diverge substantially across different architectures (Figs.~\ref{fig:bn_hist},~\ref{fig:dec_hist}).

Since all decoder activations are deterministic functions of $\mathbf{z}_S$, there exists a sub-decoder $D_S^{(k)}$ with $\mathbf{h}_S^{(k)} = D_S^{(k)}(\mathbf{z}_S)$. Assuming $D_S^{(k)}$ is $L_k$-Lipschitz, downstream mismatch satisfies:
\begin{equation}
\label{eq:lipschitz_downstream}
\|\mathbf{h}_S^{(k)} - \mathbf{h}_T^{(k)}\|
\le L_k\|\mathbf{z}_S - \mathbf{z}_T\| + \varepsilon_{\text{dec}}^{(k)},
\end{equation}
where $\varepsilon_{\text{dec}}^{(k)} = \sup_{\mathbf{z}}\|D_S^{(k)}(\mathbf{z}) - D_T^{(k)}(\mathbf{z})\|$ captures decoder-side mismatch. Under self-distillation, $\varepsilon_{\text{dec}}^{(k)}$ is small, so minimizing $\|\mathbf{z}_S - \mathbf{z}_T\|$ directly controls decoder mismatch, making explicit decoder supervision redundant. Quantization perturbations further compound across depth: modeling the $l$-th layer as $\mathbf{h}_{l+1} = \phi_l(\mathbf{h}_l) + \boldsymbol{\delta}_l$ with $\phi_l$ being $\kappa_l$-Lipschitz, the deviation at decoder layer $k$ satisfies
\begin{equation}
\label{eq:noise_accumulation}
\|h_k - h_k^{\text{FP32}}\|
\le \sum_{l=0}^{k-1} (\prod_{j=l+1}^{k-1}\kappa_j)\|{\delta}_l\|,
\end{equation}
explaining the large distribution shifts in late decoder features (Fig.~\ref{fig:dec_hist}). Imposing decoder feature-matching loss $\mathcal{L}_{\text{KD}}^{(k)} = d(\mathbf{h}_S^{(k)}, \mathbf{h}_T^{(k)})$ backpropagates through a quantization-corrupted Jacobian:
\begin{equation}
\nabla_{\mathbf{z}_S}\mathcal{L}_{\text{KD}}^{(k)}
= \left(\frac{\partial\mathbf{h}_S^{(k)}}{\partial\mathbf{z}_S}\right)^{\!\top}
\nabla_{\mathbf{h}_S^{(k)}}\, d(\mathbf{h}_S^{(k)}, \mathbf{h}_T^{(k)}),
\end{equation}
which encourages the decoder to \emph{compensate} for upstream corruption rather than \emph{repair} it at source, introducing competing gradients with $\mathcal{L}_{\text{QR}}$. Bottleneck-only distillation corrects quantization errors at their origin, yielding strong downstream alignment without decoder supervision.

\paragraph{Teacher Choice:}


Direct feature matching across heterogeneous architectures is fundamentally ill-posed, as teacher and student representations occupy incompatible feature spaces~\cite{yang2022masked}, conflating architectural mismatch with quantization induced perturbations. We therefore adopt a \emph{self-distillation} paradigm, wherein the full-precision (FP32) student serves as its own teacher, ensuring identical architecture and layer semantics so that the distillation signal exclusively targets quantization-induced deviations. As shown in Fig.~\ref{fig:bn_hist} and~\ref{fig:dec_hist}, this yields significantly closer activation alignment at both bottleneck and decoder layers, resulting in more stable optimization and stronger downstream reconstruction quality.

\subsubsection{Learnable Magnitude Reweighting}

Building on Eq.~\eqref{eq:gor_eq}, we define $r = \lambda_{\text{rec}}/\lambda_{\text{kd}}$, giving the reciprocal weighting with gradient update direction:

\begin{equation}
\nabla_{\theta_S}\mathcal{L}_{\text{QDR}}
= r\,\nabla_{\theta_S}\mathcal{L}_{\text{QR}} + \frac{1}{r}\,\nabla_{\theta_S}\mathcal{L}_{\text{KD}}
\label{eq:gor_grad}
\end{equation}
Even when losses appear balanced, gradient norms can differ substantially, causing one objective to dominate. Under QAT, quantization noise (Sec.~3.1) further destabilizes reciprocal balancing via: (i) \emph{ratio instability} (small $\lambda$ induces large reciprocal weights), (ii) \emph{sign ambiguity} (unconstrained $\lambda$ can invert objectives Fig.\ref{fig:gor_reg_params_compare}), and (iii) \emph{noisy per-iteration fluctuations} (causing oscillatory training). We propose a LMR with three improvements. First, we enforce $\lambda_{\text{rec}}=\exp(\alpha)$, $\lambda_{\text{kd}}=\exp(\beta)$, preventing sign flips and reducing sensitivity to parameter drift. Second, we track instantaneous gradient magnitudes at time $(t)$: $
g_{\text{rec}}(t)=\left\|\nabla_{\theta_S}\mathcal{L}_{\text{QR}}\right\|_2, 
g_{\text{kd}}(t)=\left\|\nabla_{\theta_S}\mathcal{L}_{\text{KD}}\right\|_2
$ and maintain EMA estimates to reduce noise: $\bar{g}_{j}(t)=\mu\,\bar{g}_{j}(t\!-\!1) + (1-\mu)\,g_{j}(t), j\in\{\text{rec},\text{kd}\}$ where $\mu\in[0,1)$ is the momentum. Third, we compute a smoothed gradient ratio to modulate the weighting:

\begin{align}
    s(t) \triangleq \sqrt{\frac{\bar{g}_{\text{kd}}(t)}{\bar{g}_{\text{rec}}(t)+\epsilon}} \qquad
    \mathcal{L}_{\text{QDR}}^{\text{LMR}} = \frac{\lambda_{\text{rec}}}{\lambda_{\text{kd}}\,s(t)}\mathcal{L}_{\text{QR}} + \frac{\lambda_{\text{kd}}}{\lambda_{\text{rec}} / s(t)}\,\mathcal{L}_{\text{KD}}
\end{align}

where $\epsilon$ prevents division by zero. Equivalently, defining an effective ratio $r_t = \frac{\lambda_{\text{rec}}}{\lambda_{\text{kd}}}$, we obtain the simplified form:

\begin{equation}
\mathcal{L}_{\text{QDR}}^{\text{LMR}} = \frac{r_t}{s(t)}\,\mathcal{L}_{\text{QR}} + \frac{s(t)}{r_t}\,\mathcal{L}_{\text{KD}}
\label{eq:gor_v2_simplified}
\end{equation}
This design preserves the simplicity of reciprocal balancing while explicitly accounting for gradient dominance and reducing oscillations under quantization noise.

\subsection{Edge-Friendly Model}

Edge deployment on resource-constrained hardware (e.g., mobile NPUs) imposes strict constraints on latency, memory, and supported operators. We design the \textbf{EFM}, a compact U-Net built exclusively from quantization-friendly operations. Please see \textbf{Supp S1} for more details. A key challenge in such architectures is the loss of localized degradation context (e.g., spatially varying illumination, rain streaks) during progressive downsampling. To address this, we propose \textbf{LDG}, a lightweight skip-connection module that preserves spatially discriminative cues with negligible parameter and compute overhead.

\paragraph{Learnable Degradation Gating: }

Naive skip connections treat all spatial locations uniformly, indiscriminately propagating both useful context and degradation-corrupted features into the decoder. LDG addresses this by making skip fusion \emph{degradation-aware} at negligible cost. At each encoder level $\ell$, we first compress $F_{\ell}^{\text{enc}}\in\mathbb{R}^{C\times H\times W}$ into a low-rank spatial descriptor via a $1{\times}1$ bottleneck:$\tilde{F}_{\ell} = \varphi(F_{\ell}^{\text{enc}}) \in \mathbb{R}^{\frac{C}{r}\times H\times W},$ where $\varphi = \text{ReLU}(\text{BN}(\text{Conv}_{1\times1}(\cdot)))$ and $r{=}8$. A single-channel degradation importance map is then predicted via a $3{\times}3$ convolution followed by sigmoid activation:


\begin{equation}
    G_{\ell} = \sigma\!\left(\text{proj}(\tilde{F}_{\ell})\right) \in \mathbb{R}^{1\times H\times W},
\end{equation}

where each spatial entry of $G_{\ell}$ encodes a per-pixel \emph{degradation importance score}, indicating how strongly the corresponding encoder feature should influence the decoder. 






Rather than pure multiplicative attention, which can induce quantization dead-zones in lightly degraded regions, we apply a \emph{learnable residual gating} ($S_{\ell}$) and fuse it directly into the decoder features ($F_{\ell}^{\text{dec}}$) as follows:

\begin{equation}
\label{eq:ldg_fusion}
S_{\ell} = F_{\ell}^{\text{enc}} \odot \left(\alpha_{\text{deg}} + G_{\ell}\right), \quad \text{and} \quad F_{\ell}^{\text{dec}} = \text{Up}\!\left(F_{\ell+1}^{\text{dec}}\right) + \alpha_{\ell}\, S_{\ell}.
\end{equation}

Here, $\odot$ denotes channel-broadcast multiplication, and $[\alpha_{\ell},\, \alpha_{\text{deg}}]$ are learnable scalars initialized to $1$. The additive offset $\alpha_{\text{deg}}$ ensures non-trivial gradient flow and prevents Int8 activation dead-zones during QAT. Meanwhile, scalar fusion via $\alpha_{\ell}$ dynamically weights the gated skip connection, avoiding the severe channel-width expansion incurred by standard concatenation.

\vspace{-0.2cm}
\section{Experiments}
 \paragraph{Dataset and Methods.} We evaluate our proposed framework across four common image restoration tasks: image dehazing on SOTS \cite{li2018benchmarking}, deraining on Rain100H \cite{yang2017deep}, real-world denoising on SIDD \cite{abdelhamed2018high}, and low-light enhancement on LOL-v1 \cite{wei2018deep}. To comprehensively assess our method, we establish two sets of baselines. First, for architectural comparison, we benchmark our EFM against SOTA full-precision (FP32) networks. These include CNN-based models (U-Net\cite{ronneberger2015u}, DnCNN \cite{zhang2017beyond}, LEDNet \cite{zhou2022lednet}, LPIENet \cite{conde2023perceptual}, DarkIR \cite{feijoo2025darkir}, Optimized LLIE\cite{a2024learning}, ELIR\cite{cohen2025efficient}) and Transformer-based architectures (Uformer\cite{wang2022uformer}, Restormer\cite{zamir2022restormer}, MIRNet-v2\cite{zamir2022learning}). Second, to evaluate our QDR framework, we compare our DFD against standard quantization and KD for image restoration under QAT settings. These include PTQ, QAT, KD, and recent distillation methods such as FAKD \cite{yuan2024fakd}, DCKD \cite{zhou2025dynamic}, and SLKD \cite{zhang2025knowledge}. For fair comparison, we use our fp32 model as the teacher for all KD methods. In addition, we used retinex\cite{cai2023retinexformer} and mirnetv2\cite{Zamir2022MIRNetv2} to evaluate the practicability of heterogeneous teachers in IR. 
\paragraph{Implementation Details.} Our network is trained from scratch using a combination of $L_1$ and SSIM losses. Optimization is performed using the Adam optimizer with an initial learning rate of $1 \times 10^{-5}$. During training, we extract $256 \times 256$ patches with a batch size of 32. The framework is implemented in PyTorch \cite{paszke2019pytorch}, and quantization is executed using the NVIDIA ModelOpt toolkit.We quantize both weights and activations to Int8 precision to assess practical deployment efficiency and restoration performance.  To evaluate real-world edge deployment, inference speeds (FPS) are benchmarked on both a high-end NVIDIA RTX 3090 GPU and a resource-constrained Jetson Orin edge device \cite{karumbunathan2022nvidia}.

\subsection{Comparison}
\subsubsection{Comparison with Restoration Method}


\paragraph{Quantitative Evaluation.} 
\begin{table*}[!htb]
\centering
\caption{Quantitative comparison with state-of-the-art methods. \best{Red/Bold} and \second{Blue/Underlined} indicates the best and the second best among CNNs respectively.} 
\label{tab:main_results}


\resizebox{\textwidth}{!}{%
\begin{tabular}{llccccccccccccccccccccc}
\toprule
\multirow{2}{*}{\textbf{Type}} & \multirow{2}{*}{\textbf{Method}} & \multicolumn{4}{c}{\textbf{Complexity}} & \multicolumn{3}{c}{\textbf{Denoise (SIDD)}} & \multicolumn{3}{c}{\textbf{Lowlight (LoL-v1)}} & \multicolumn{3}{c}{\textbf{Derain (Rain100H)}} & \multicolumn{3}{c}{\textbf{Dehaze (SOTS)}} & \multicolumn{3}{c}{\textbf{Average}} \\
\cmidrule(lr){3-6} \cmidrule(lr){7-9} \cmidrule(lr){10-12} \cmidrule(lr){13-15} \cmidrule(lr){16-18} \cmidrule(lr){19-21}
 & & \textbf{Flops} & \textbf{Params} & \textbf{FPS\textsubscript{3090}} & \textbf{FPS\textsubscript{ORIN}} & \textbf{PSNR}~$\uparrow$ & \textbf{SSIM}~$\uparrow$ & \textbf{LPIPS}~$\downarrow$ & \textbf{PSNR}~$\uparrow$ & \textbf{SSIM}~$\uparrow$ & \textbf{LPIPS}~$\downarrow$ & \textbf{PSNR}~$\uparrow$ & \textbf{SSIM}~$\uparrow$ & \textbf{LPIPS}~$\downarrow$ & \textbf{PSNR}~$\uparrow$ & \textbf{SSIM}~$\uparrow$ & \textbf{LPIPS}~$\downarrow$ & \textbf{PSNR}~$\uparrow$ & \textbf{SSIM}~$\uparrow$ & \textbf{LPIPS}~$\downarrow$ \\
\midrule

\multirow{3}{*}{\rotatebox{90}{Trans.}} 
 & Uformer \cite{wang2022uformer} & 10.87 & 5.29 & 52 & - & 37.20 & 0.879 & 0.210 & 22.83 & 0.870 & 0.108 & 26.29 & 0.829 & 0.118 & 25.98 & 0.948 & 0.027 & 28.08 & 0.882 & 0.116 \\
 & Restormer \cite{zamir2022restormer} & 141.24 & 26.13 & 9 & - & 40.02 & 0.960 & 0.149 & 20.41 & 0.806 & 0.128 & 31.46 & 0.904 & 0.081 & 30.87 & 0.969 & 0.018 & 30.69 & 0.910 & 0.094 \\
 & MIRNetv2 \cite{Zamir2022MIRNetv2} & 140.92 & 5.18 & 17 & 6 & 39.84 & 0.959 & 0.152 & 24.74 & 0.851 & 0.071 & 26.35 & 0.824 & 0.115 & 26.26 & 0.956 & 0.035 & 29.30 & 0.898 & 0.093 \\
\midrule

\multirow{8}{*}{\rotatebox{90}{CNN}} 
 & U-Net \cite{ronneberger2015u} & 15.53 & 9.50 & 207 & \second{144} & 27.33 & 0.494 & 0.559 & 17.18 & 0.722 & 0.305 & 19.97 & 0.598 & 0.375 & 24.41 & 0.932 & 0.046 & 22.22 & 0.687 & 0.321 \\
 & DnCNN \cite{zhang2017beyond} & 36.73 & 0.67 & 172 & 89 & 31.53 & 0.933 & 0.249 & 18.67 & 0.732 & 0.246 & 24.28 & 0.796 & 0.176 & 25.18 & 0.928 & 0.060 & 24.92 & 0.847 & 0.183 \\
 & LEDNet \cite{zhou2022lednet} & 38.66 & 7.41 & 132 & - & 27.65 & 0.798 & 0.351 & 19.23 & 0.700 & 0.222 & 23.12 & 0.771 & 0.259 & 27.01 & 0.912 & 0.042 & 24.25 & 0.795 & 0.219 \\
 & LPIENet \cite{conde2023perceptual} & \best{4.29} & \best{0.45} & \best{246} & \best{337} & 26.55 & 0.518 & 0.534 & 18.95 & 0.781 & 0.233 & 24.16 & 0.764 & 0.191 & 26.26 & 0.956 & 0.036 & 23.98 & 0.755 & 0.249 \\
 & DarkIR \cite{feijoo2025darkir} & 27.19 & 12.96 & 79 & - & \second{36.86} & \second{0.875} & \second{0.236} & \second{23.11} & 0.856 & 0.088 & 26.11 & 0.832 & 0.115 & 27.15 & 0.947 & 0.028 & \second{28.31} & \second{0.878} & 0.117 \\
 & Opt-LLIE \cite{a2024learning} & 23.10 & 4.47 & 195 & 136 & 36.00 & 0.864 & 0.239 & 20.10 & 0.778 & 0.225 & \second{28.06} & \second{0.842} & 0.159 & 25.58 & 0.896 & 0.068 & 27.44 & 0.845 & 0.173 \\
 & ELIR \cite{cohen2025efficient} & 409.17 & 26.89 & 35 & - & 29.96 & 0.853 & 0.136 & 14.75 & 0.531 & 0.531 & 21.16 & 0.616 & 0.324 & 22.70 & 0.807 & 0.108 & 22.14 & 0.702 & 0.275 \\
 
\rowcolor{rowgray}
 & \textbf{Ours} & \second{19.90} & \second{4.01} & \second{216} & 139 & \begin{tabular}[c]{@{}c@{}}\best{37.22}\\{\scriptsize\textcolor{green!60!black}{+0.36}}\end{tabular} & \begin{tabular}[c]{@{}c@{}}\best{0.877}\\{\scriptsize\textcolor{green!60!black}{+0.002}}\end{tabular} & \begin{tabular}[c]{@{}c@{}}\best{0.227}\\{\scriptsize\textcolor{green!60!black}{-0.009}}\end{tabular} & \begin{tabular}[c]{@{}c@{}}\best{23.19}\\{\scriptsize\textcolor{green!60!black}{+0.08}}\end{tabular} & \begin{tabular}[c]{@{}c@{}}\best{0.878}\\{\scriptsize\textcolor{green!60!black}{+0.022}}\end{tabular} & \begin{tabular}[c]{@{}c@{}}\best{0.085}\\{\scriptsize\textcolor{green!60!black}{-0.003}}\end{tabular} & \begin{tabular}[c]{@{}c@{}}\best{30.03}\\{\scriptsize\textcolor{green!60!black}{+1.97}}\end{tabular} & \begin{tabular}[c]{@{}c@{}}\best{0.886}\\{\scriptsize\textcolor{green!60!black}{+0.044}}\end{tabular} & \begin{tabular}[c]{@{}c@{}}\best{0.111}\\{\scriptsize\textcolor{green!60!black}{-0.004}}\end{tabular} & \begin{tabular}[c]{@{}c@{}}\best{28.10}\\{\scriptsize\textcolor{green!60!black}{+0.95}}\end{tabular} & \begin{tabular}[c]{@{}c@{}}\best{0.961}\\{\scriptsize\textcolor{green!60!black}{+0.014}}\end{tabular} & \begin{tabular}[c]{@{}c@{}}\best{0.018}\\{\scriptsize\textcolor{green!60!black}{+0.000}}\end{tabular} & \begin{tabular}[c]{@{}c@{}}\best{29.64}\\{\scriptsize\textcolor{green!60!black}{+1.33}}\end{tabular} & \begin{tabular}[c]{@{}c@{}}\best{0.901}\\{\scriptsize\textcolor{green!60!black}{+0.023}}\end{tabular} & \begin{tabular}[c]{@{}c@{}}\best{0.110}\\{\scriptsize\textcolor{green!60!black}{-0.007}}\end{tabular} \\
\bottomrule
\end{tabular}%
}
\end{table*}

\vspace{-0.cm}
We benchmark EFM against SOTA methods across four tasks to establish it as a baseline for edge-deployable restoration in full precision settings (Tab.~\ref{tab:main_results}). Among CNN-based architectures, our approach achieves the best \emph{average} performance (29.64\,dB PSNR / 0.901 SSIM). It consistently outperforms strong baselines across all tasks, yielding gains of up to +1.97\,dB on Rain100H \cite{yang2017deep} and +0.95\,dB on SOTS \cite{li2018benchmarking}. Overall, our method achieves the best trade-off between accuracy and efficiency. Transformer-based models (e.g., Restormer) can achieve superior restoration performance on a few specific tasks; however, they are computationally expensive and rely on unsupported edge operations. In contrast, our model is lightweight (4.01M params, 19.9 GFLOPs) and delivers high throughput, achieving \textbf{216 FPS} on an NVIDIA RTX 3090 and \textbf{139 FPS} (FP32) on a Jetson Orin.

\begin{figure}[!htb]
    \centering
    \includegraphics[width=\linewidth, height=5.6cm]{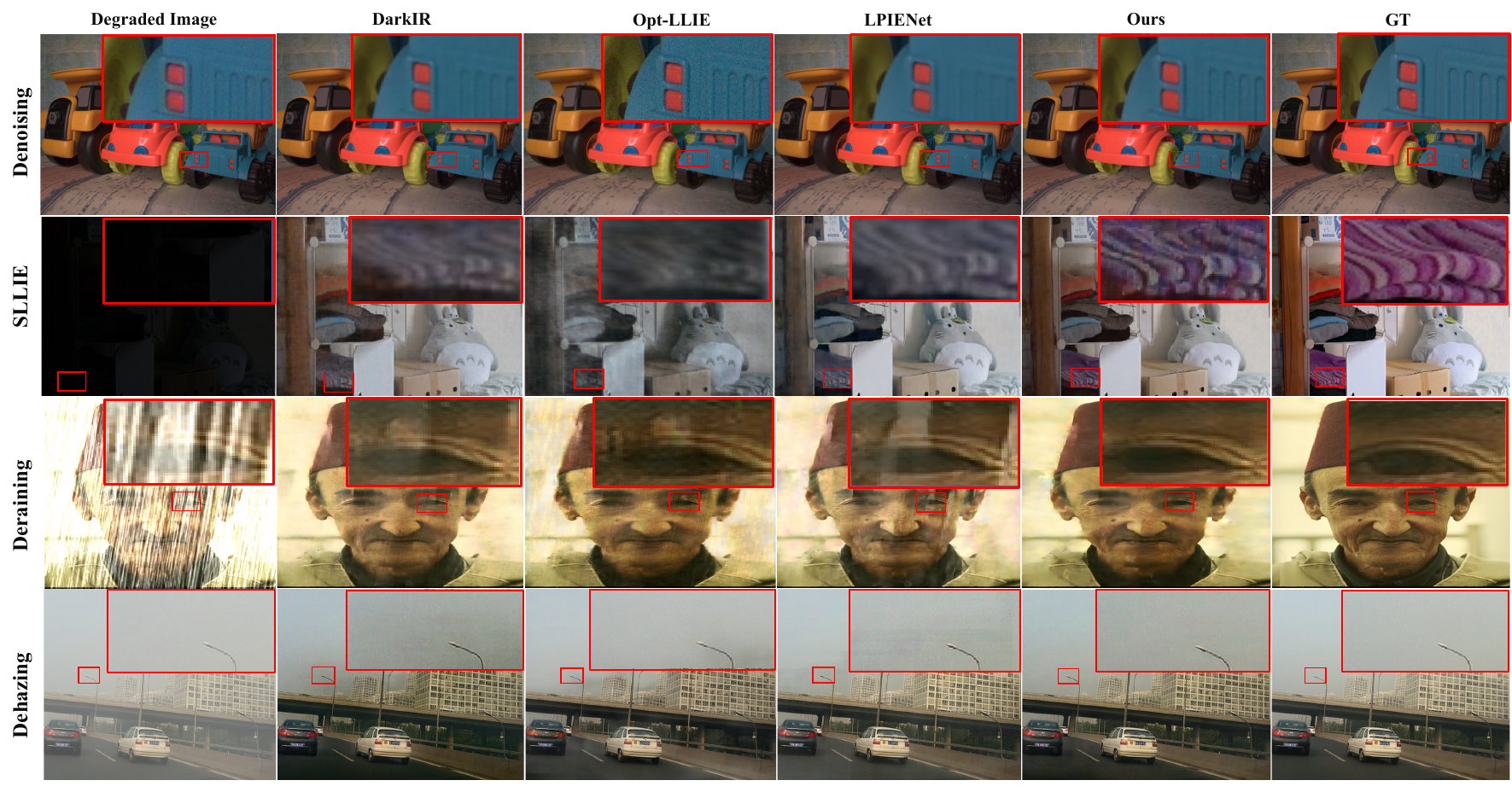}
    \caption{Qualitative results on four restoration tasks: denoising, LLIE, deraining, and dehazing on full-precision settings.}
    \label{fig:vis_comp}
    \vspace{-0.1cm}
\end{figure}

\paragraph{Qualitative Evaluation.}
Fig.~\ref{fig:vis_comp} presents visual comparisons in the full-precision setting. Consistent with the quantitative results, our method suppresses artifacts more effectively than DarkIR \cite{feijoo2025darkir}, Opt-LLIE \cite{a2024learning}, and LPIENet \cite{conde2023perceptual}. As highlighted in the red boxes, our approach better preserves high-frequency details and fine textures, avoiding the over-smoothing common in competing lightweight networks.

\subsubsection{Comparison with SOTA KD under QAT}
\paragraph{Quantitative Evaluation.}

As shown in Tab.~\ref{tab:quantization_results}, our approach significantly outperforms standard quantization baselines. Combining KD with QAT yields better results than QAT alone; these methods still fail to match the full-precision (FP32) upper bound. In contrast, DFD achieves SOTA Int8 performance with \textbf{28.60\,dB} PSNR, surpassing the strongest baseline (FAKD \cite{yuan2024fakd}) by \textbf{+0.67\,dB} and yields the lowest perceptual error (LPIPS \cite{zhang2018unreasonable}).
\begin{table*}[!htb]
\centering
\caption{Quantitative comparison of INT8 quantization methods. We compare various quantization approaches on four restoration tasks. \best{Red/Bold} indicates the best while \second{Blue/Underlined} indicates the second best.}
\label{tab:quantization_results}

\setlength{\tabcolsep}{2.8pt}
\renewcommand{\arraystretch}{1.25}

\resizebox{\textwidth}{!}{%
\begin{tabular}{lccccccccccccccc}
\toprule
\multirow{2}{*}{\textbf{Method}} & \multicolumn{3}{c}{\textbf{Denoise (SSID)}} & \multicolumn{3}{c}{\textbf{Lowlight (LoL-v1)}} & \multicolumn{3}{c}{\textbf{Derain (Rain100H)}} & \multicolumn{3}{c}{\textbf{Dehaze (SOTS)}} & \multicolumn{3}{c}{\textbf{Average}} \\
\cmidrule(lr){2-4} \cmidrule(lr){5-7} \cmidrule(lr){8-10} \cmidrule(lr){11-13} \cmidrule(lr){14-16}
 & \textbf{PSNR}~$\uparrow$ & \textbf{SSIM}~$\uparrow$ & \textbf{LPIPS}~$\downarrow$ & \textbf{PSNR}~$\uparrow$ & \textbf{SSIM}~$\uparrow$ & \textbf{LPIPS}~$\downarrow$ & \textbf{PSNR}~$\uparrow$ & \textbf{SSIM}~$\uparrow$ & \textbf{LPIPS}~$\downarrow$ & \textbf{PSNR}~$\uparrow$ & \textbf{SSIM}~$\uparrow$ & \textbf{LPIPS}~$\downarrow$ & \textbf{PSNR}~$\uparrow$ & \textbf{SSIM}~$\uparrow$ & \textbf{LPIPS}~$\downarrow$ \\
\midrule

\rowcolor{gray!15}
\textit{FP32 (Reference)} & \textit{37.22} & \textit{0.877} & \textit{0.227} & \textit{22.43} & \textit{0.799} & \textit{0.174} & \textit{30.03} & \textit{0.886} & \textit{0.111} & \textit{28.10} & \textit{0.961} & \textit{0.018}  & \textit{29.64} & \textit{0.900} & \textit{0.110} \\
\midrule

PTQ & 29.31 & 0.584 & 0.491 & 18.86 & 0.579 & 0.398 & 28.10 & 0.814 & 0.165 & 26.85 & 0.919 & 0.037 & 25.78 & 0.724 & 0.273 \\
QAT & 33.09 & 0.729 & 0.491 & 21.76 & 0.749 & 0.233 & 28.36 & 0.823 & 0.158 & 27.19 & 0.926 & 0.033 & 27.60 & 0.809 & 0.222 \\
QAT + KD \cite{hinton2015distilling}& \second{33.58} & \second{0.734} & \second{0.358} & 21.99 & 0.758 & 0.226 & 28.47 & 0.828 & \second{0.154} & 27.51 & 0.928 & 0.030 & 27.89 & \second{0.817} & \second{0.183} \\
QAT + FAKD\cite{yuan2024fakd} & 33.31 & 0.709 & 0.491 & \second{22.10} & \second{0.762} & \second{0.223} & 28.44 & 0.827 & 0.155 & 27.86 & 0.931 & 0.030 & \second{27.93} & 0.816 & 0.212 \\
QAT + SLKD \cite{zhang2025knowledge} & 31.72 & 0.695 & 0.452 & 21.78 & 0.750 & 0.232 & 28.41 & 0.825 & 0.156 & \second{28.09} & \second{0.934} & \second{0.024} & 27.50 & 0.797 & 0.210 \\
QAT + DCKD \cite{zhou2025dynamic} & 25.66 & 0.460 & 0.625 & 14.64 & 0.422 & 0.626 & \second{28.61} & \second{0.833} & 0.151 & 26.68 & 0.917 & 0.039 & 23.90 & 0.658 & 0.360 \\

\midrule
\rowcolor{rowgray}
\textbf{QDR (Ours)} & \begin{tabular}[c]{@{}c@{}}\best{34.06}\\{\scriptsize\textcolor{green!60!black}{+0.48}}\end{tabular} & \begin{tabular}[c]{@{}c@{}}\best{0.749}\\{\scriptsize\textcolor{green!60!black}{+0.015}}\end{tabular} & \begin{tabular}[c]{@{}c@{}}\best{0.334}\\{\scriptsize\textcolor{green!60!black}{-0.024}}\end{tabular} & \begin{tabular}[c]{@{}c@{}}\best{22.36}\\{\scriptsize\textcolor{green!60!black}{+0.26}}\end{tabular} & \begin{tabular}[c]{@{}c@{}}\best{0.782}\\{\scriptsize\textcolor{green!60!black}{-0.020}}\end{tabular} & \begin{tabular}[c]{@{}c@{}}\best{0.198}\\{\scriptsize\textcolor{green!60!black}{-0.025}}\end{tabular} & \begin{tabular}[c]{@{}c@{}}\best{28.95}\\{\scriptsize\textcolor{green!60!black}{+0.34}}\end{tabular} & \begin{tabular}[c]{@{}c@{}}\best{0.833}\\{\scriptsize\textcolor{green!60!black}{+0.000}}\end{tabular} & \begin{tabular}[c]{@{}c@{}}\best{0.147}\\{\scriptsize\textcolor{green!60!black}{-0.004}}\end{tabular} & \begin{tabular}[c]{@{}c@{}}\best{28.72}\\{\scriptsize\textcolor{green!60!black}{+0.63}}\end{tabular} & \begin{tabular}[c]{@{}c@{}}\best{0.962}\\{\scriptsize\textcolor{green!60!black}{+0.028}}\end{tabular} & \begin{tabular}[c]{@{}c@{}}\best{0.016}\\{\scriptsize\textcolor{green!60!black}{-0.008}}\end{tabular} & \begin{tabular}[c]{@{}c@{}}\best{28.60}\\{\scriptsize\textcolor{green!60!black}{+0.67}}\end{tabular} & \begin{tabular}[c]{@{}c@{}}\best{0.832}\\{\scriptsize\textcolor{green!60!black}{+0.015}}\end{tabular} & \begin{tabular}[c]{@{}c@{}}\best{0.167}\\{\scriptsize\textcolor{green!60!black}{-0.016}}\end{tabular} \\
\bottomrule
\end{tabular}%
}
\end{table*}

\vspace{-0.8cm}

\paragraph{Qualitative Evaluation.}

\begin{figure}[!htb]
    \centering
    \includegraphics[width=\linewidth, height=5.6cm]{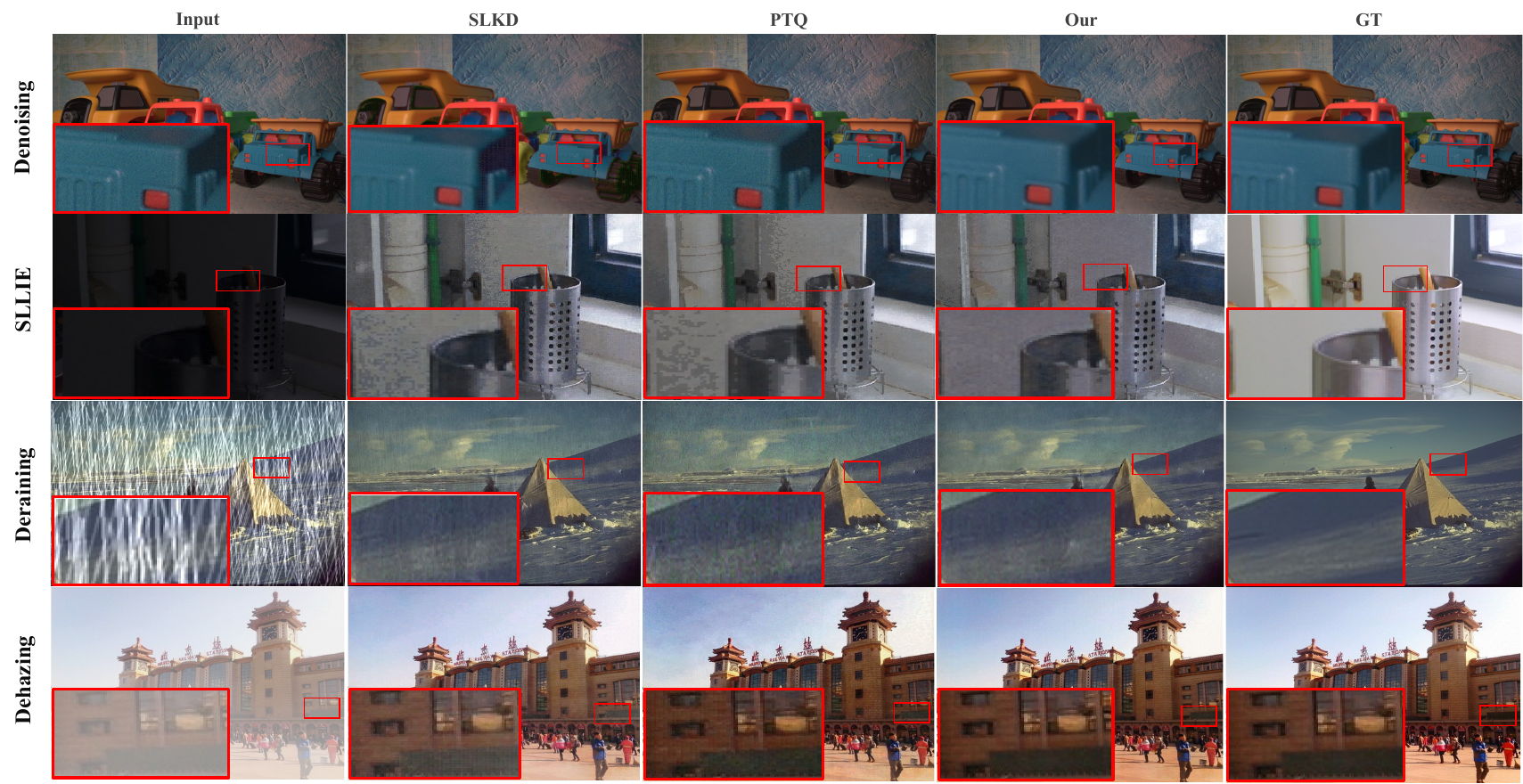}
    \caption{Qualitative results for quantized image restoration (8-bit): denoising, LLIE, deraining, and dehazing.}
    \label{fig:results_kd}
    \vspace{-0.3cm}
\end{figure}

Fig.~\ref{fig:results_kd} further confirms the practicability of DFD with visual comparison. Methods like QAT+SLKD\cite{zhang2025knowledge} improve stability over standard PTQ. However,  it tends to over-smooth complex regions or leave residual noise (see red boxes). Our method, DFD, restores sharper fine structures and cleaner edges, aligning with the quantitative gains reported in Tab.~\ref{tab:quantization_results}.

\subsection{Generalization}

Beyond standard 8-bit weight and activation (W/A) quantization, we push the limits of our QDR framework by extending it to extremely low-bit regimes (2-bit and 4-bit). We evaluate performance across three challenging restoration tasks: low-light enhancement, dehazing, and deraining. Furthermore, to demonstrate the plug-and-play generalizability of our QDR, we apply it to an existing SOTA architecture, Opt-LLIE, chosen for its strong trade-off between performance and edge-deployment efficiency. As shown in Tab.~\ref{tab:bitwidth_comparison}, our QDR strategy consistently yields significant performance gains over PTQ across all bit-widths and tasks. On average across both architectures, our method improves restoration quality by 0.61 dB at 2-bit, 5.47 dB at 4-bit, and 0.74 dB at 8-bit quantization compared to the PTQ baseline. Please refer to \textbf{Supp. S3} for more results.
\begin{table*}[!htb]
\centering
\caption{Quantitative comparison of proposed QDR vs. PTQ across 2-, 4-, and 8-bit widths. Our method consistently achieves higher \textbf{PSNR}~($\uparrow$) on all tasks.}
\label{tab:bitwidth_comparison}

\setlength{\tabcolsep}{3.5pt}
\renewcommand{\arraystretch}{1.25}

\resizebox{\textwidth}{!}{%
\begin{tabular}{lccccccccccccccccccccc}
\toprule
\multirow{3}{*}{\textbf{Model}} & \multicolumn{7}{c}{\textbf{Lowlight (LoL-v1)}} & \multicolumn{7}{c}{\textbf{Dehazing (SOTS)}} & \multicolumn{7}{c}{\textbf{Deraining (Rain100H)}} \\
\cmidrule(lr){2-8} \cmidrule(lr){9-15} \cmidrule(lr){16-22}
 & \multicolumn{2}{c}{\textbf{2-bit}} & \multicolumn{2}{c}{\textbf{4-bit}} & \multicolumn{2}{c}{\textbf{8-bit}} & \textbf{FP32} & \multicolumn{2}{c}{\textbf{2-bit}} & \multicolumn{2}{c}{\textbf{4-bit}} & \multicolumn{2}{c}{\textbf{8-bit}} & \textbf{FP32} & \multicolumn{2}{c}{\textbf{2-bit}} & \multicolumn{2}{c}{\textbf{4-bit}} & \multicolumn{2}{c}{\textbf{8-bit}} & \textbf{FP32} \\
\cmidrule(lr){2-3} \cmidrule(lr){4-5} \cmidrule(lr){6-7} \cmidrule(lr){9-10} \cmidrule(lr){11-12} \cmidrule(lr){13-14} \cmidrule(lr){16-17} \cmidrule(lr){18-19} \cmidrule(lr){20-21}
 & \textbf{PTQ} & \textbf{DFD} & \textbf{PTQ} & \textbf{DFD} & \textbf{PTQ} & \textbf{DFD} & & \textbf{PTQ} & \textbf{DFD} & \textbf{PTQ} & \textbf{DFD} & \textbf{PTQ} & \textbf{DFD} & & \textbf{PTQ} & \textbf{DFD} & \textbf{PTQ} & \textbf{DFD} & \textbf{PTQ} & \textbf{DFD} & \\
\midrule

Opt-LLIE \cite{a2024learning} & 9.30 & \begin{tabular}[c]{@{}c@{}}9.80\\{\scriptsize\textcolor{green!60!black}{+0.50}}\end{tabular} & 10.64 & \begin{tabular}[c]{@{}c@{}}18.20\\{\scriptsize\textcolor{green!60!black}{+7.56}}\end{tabular} & 20.04 & \begin{tabular}[c]{@{}c@{}}21.20\\{\scriptsize\textcolor{green!60!black}{+1.16}}\end{tabular} & 20.10 & 14.01 & \begin{tabular}[c]{@{}c@{}}14.30\\{\scriptsize\textcolor{green!60!black}{+0.29}}\end{tabular} & 19.27 & \begin{tabular}[c]{@{}c@{}}22.69\\{\scriptsize\textcolor{green!60!black}{+3.42}}\end{tabular} & 25.10 & \begin{tabular}[c]{@{}c@{}}25.47\\{\scriptsize\textcolor{green!60!black}{+0.37}}\end{tabular} & 25.58 & 11.93 & \begin{tabular}[c]{@{}c@{}}10.70\\{\scriptsize\textcolor{red}{-1.23}}\end{tabular} & 18.22 & \begin{tabular}[c]{@{}c@{}}22.52\\{\scriptsize\textcolor{green!60!black}{+4.30}}\end{tabular} & 26.97 & \begin{tabular}[c]{@{}c@{}}27.18\\{\scriptsize\textcolor{green!60!black}{+0.21}}\end{tabular} & 28.06 \\

\rowcolor{rowgray}
\textbf{Ours} & \textbf{5.82} & \begin{tabular}[c]{@{}c@{}}\textbf{7.86}\\{\scriptsize\textcolor{green!60!black}{+2.04}}\end{tabular} & \textbf{9.91} & \begin{tabular}[c]{@{}c@{}}\textbf{13.47}\\{\scriptsize\textcolor{green!60!black}{+3.56}}\end{tabular} & \textbf{21.10} & \begin{tabular}[c]{@{}c@{}}\textbf{22.36}\\{\scriptsize\textcolor{green!60!black}{+1.26}}\end{tabular} & \textbf{22.43} & \textbf{17.71} & \begin{tabular}[c]{@{}c@{}}\textbf{18.20}\\{\scriptsize\textcolor{green!60!black}{+0.49}}\end{tabular} & \textbf{16.97} & \begin{tabular}[c]{@{}c@{}}\textbf{23.80}\\{\scriptsize\textcolor{green!60!black}{+6.83}}\end{tabular} & \textbf{26.50} & \begin{tabular}[c]{@{}c@{}}\textbf{28.72}\\{\scriptsize\textcolor{green!60!black}{+1.92}}\end{tabular} & \textbf{28.10} & \textbf{14.34} & \begin{tabular}[c]{@{}c@{}}\textbf{15.90}\\{\scriptsize\textcolor{green!60!black}{+1.56}}\end{tabular} & \textbf{11.34} & \begin{tabular}[c]{@{}c@{}}\textbf{18.47}\\{\scriptsize\textcolor{green!60!black}{+7.13}}\end{tabular} & \textbf{28.10} & \begin{tabular}[c]{@{}c@{}}\textbf{28.95}\\{\scriptsize\textcolor{green!60!black}{+0.85}}\end{tabular} & \textbf{30.03} \\

\bottomrule
\end{tabular}%
}
\end{table*}

\vspace{-0.5cm}

\subsection{Analysis}
\subsubsection{Decoder-free distillation}
\begin{wrapfigure}{r}{0.5\linewidth}
\vspace{-.9cm}
  \centering
  \includegraphics[width=\linewidth]{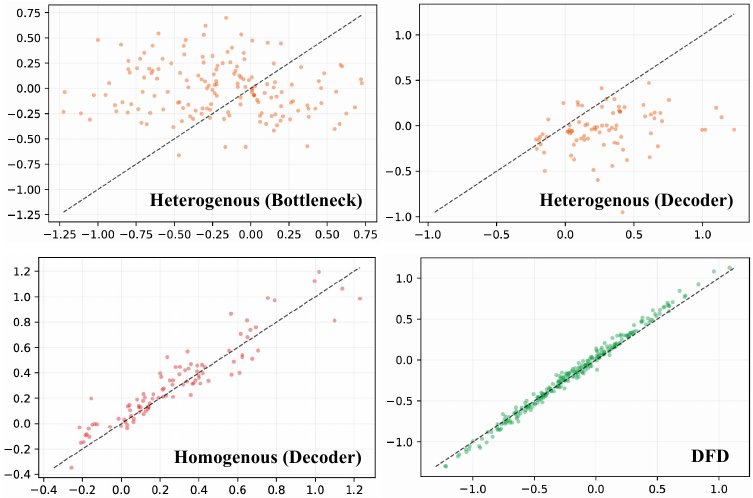}
    \caption{ Feature distribution alignment between different teachers and positions with the target FP32 Model.}
    \label{fig:alignment}
  \vspace{-.8cm}
\end{wrapfigure} 

We study two practical factors for effective QDR: \textbf{(i) teacher selection} and \textbf{(ii) distillation location}. As shown in Tab.~\ref{tab:dfd_analysis}, heterogeneous teachers (e.g., MIRNet-v2, RetinexFormer) degrade the performance of the quantized model below the QAT baseline (21.76\,dB) regardless of the distillation location. Notably, the bottleneck distillation consistently outperforms decoder distillation even for these heterogeneous models (e.g., 20.50\,dB vs.\ 19.90\,dB for Retinexformer). Our homogeneous (FP32) teacher with bottleneck distillation significantly outperforms the decoder (22.36\,dB) and other distillation positions and teachers. The bottleneck features offer compact, semantically rich representations that provide robust supervision against quantization noise. Fig.~\ref{fig:alignment} (bottom row) confirms this: bottleneck alignment yields a near-perfect diagonal correlation, whereas decoder alignment remains scattered.

\begin{table}[!htb]
\centering
\begin{minipage}{0.67\textwidth}
    \centering
\caption{Comparison of different teacher and distillation positions under QAT settings on LoLv1. }
\label{tab:dfd_analysis}


\scalebox{0.7}{%
\begin{tabular}{l c cc cc cc}
\toprule
\multirow{2}{*}{\textbf{Metric}} & \textbf{Baseline} & \multicolumn{2}{c}{\textbf{Retinexformer \cite{cai2023retinexformer}}} & \multicolumn{2}{c}{\textbf{MIRNet-v2 \cite{Zamir2022MIRNetv2}}} & \multicolumn{2}{c}{\textbf{EFM (Ours)}} \\
\cmidrule(lr){3-4} \cmidrule(lr){5-6} \cmidrule(lr){7-8}
 & (QAT) & Decoder & Bottleneck & Decoder & Bottleneck & Decoder & \cellcolor{rowgray}\textbf{Bottleneck} \\ 
\midrule
\textbf{PSNR}~$\uparrow$ & 21.76 & 
\begin{tabular}[c]{@{}c@{}}19.90\\{\scriptsize\textcolor{red}{-1.86}}\end{tabular} & 
\begin{tabular}[c]{@{}c@{}}20.50\\{\scriptsize\textcolor{red}{-1.26}}\end{tabular} & 
\begin{tabular}[c]{@{}c@{}}19.24\\{\scriptsize\textcolor{red}{-2.52}}\end{tabular} & 
\begin{tabular}[c]{@{}c@{}}20.37\\{\scriptsize\textcolor{red}{-1.39}}\end{tabular} & 
\begin{tabular}[c]{@{}c@{}}20.05\\{\scriptsize\textcolor{red}{-1.71}}\end{tabular} & 
\cellcolor{rowgray}\begin{tabular}[c]{@{}c@{}}\best{22.36}\\{\scriptsize\textcolor{green!60!black}{+0.60}}\end{tabular} \\
\bottomrule
\end{tabular}%
}

\end{minipage}
\hfill
\begin{minipage}{0.30\textwidth}
    \centering
\centering
\caption{Comparison of magnitude reweighting.}
\label{tab:reweighting}

\scalebox{.63}{%
\begin{tabular}{lccc}
\toprule
\textbf{Task} & \textbf{w/o Reb.} & \textbf{GOR~\cite{rehman2025punching}} & \cellcolor{rowgray}\textbf{Ours} \\ 
\midrule

Deraining & 28.47 & 28.62 & 
\cellcolor{rowgray}\begin{tabular}[c]{@{}c@{}}\best{28.95}\\{\scriptsize\textcolor{green!60!black}{+0.33}}\end{tabular} \\ 

Lowlight & 21.99 & 20.87 & 
\cellcolor{rowgray}\begin{tabular}[c]{@{}c@{}}\best{22.36}\\{\scriptsize\textcolor{green!60!black}{+1.49}}\end{tabular} \\

\bottomrule
\end{tabular}%
}


\end{minipage}
\vspace{-0.3cm}
\end{table}

\subsubsection{Learnable Magnitude Reweighting}
\begin{wrapfigure}{r}{0.4\linewidth}
  \centering
  \includegraphics[width=\linewidth]{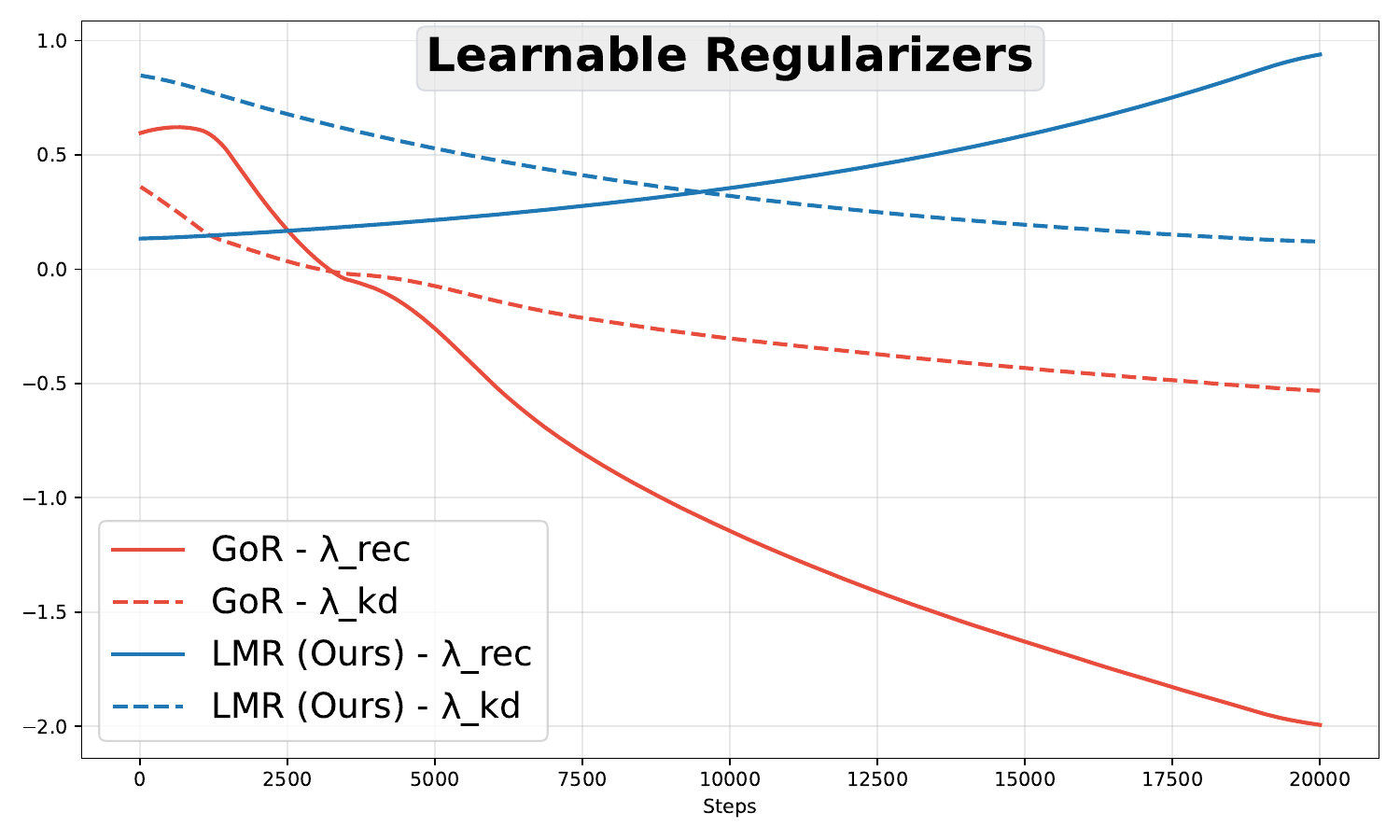}
    \caption{Comparison of GoR\cite{rehman2025punching} with LMR during a training run.}
  \vspace{-.8cm}
  \label{fig:gor_reg_params_compare}
\end{wrapfigure} 

Furthermore, we compare our proposed LMR against the existing QAT-KD reweighting method (i.e., GoR). Under the DFD framework (Tab.~\ref{tab:reweighting}), LMR consistently outperforms GoR \cite{rehman2025punching}, demonstrating robust adaptability through substantial gains in both low-light (+1.49 dB) and deraining (+0.33 dB) tasks. Fig. \ref{fig:gor_reg_params_compare} compares how trainable parameters evolve as the training progresses. It is evident that under extreme cases GoR\cite{rehman2025punching} inverts the magnitudes of the two losses which destabilizes the training.

\subsubsection{Learnable Degradation Gating}
\begin{figure}[!htb]
    \centering
    
    \begin{minipage}{0.34\textwidth}
        \centering



\centering


\scalebox{0.64}{%
\begin{tabular}{l c c c}
\toprule
\textbf{Comp.} & \textbf{Baseline} & \textbf{+ DM} & \cellcolor{rowgray}\textbf{+ LG (Ours)} \\
\midrule
\textbf{SC} & \checkmark & \checkmark & \cellcolor{rowgray}\checkmark \\
\textbf{DM} & & \checkmark & \cellcolor{rowgray}\checkmark \\
\textbf{LG} & & & \cellcolor{rowgray}\checkmark \\
\midrule
\textbf{PSNR (dB)} & 27.11 & 
\begin{tabular}[c]{@{}c@{}}28.73\\{\scriptsize\textcolor{green!60!black}{+1.62}}\end{tabular} & 
\cellcolor{rowgray}\begin{tabular}[c]{@{}c@{}}\textbf{30.03}\\{\scriptsize\textcolor{green!60!black}{+1.30}}\end{tabular} \\
\bottomrule
\end{tabular}%
}

         \captionof{table}{Analysis of LDG.}
        \label{tab:degmap_abl}
    \end{minipage}
    \hspace{0.13cm}
    \begin{minipage}{0.6\textwidth}
        \centering
        \includegraphics[width=7.4cm, height=2.051cm]{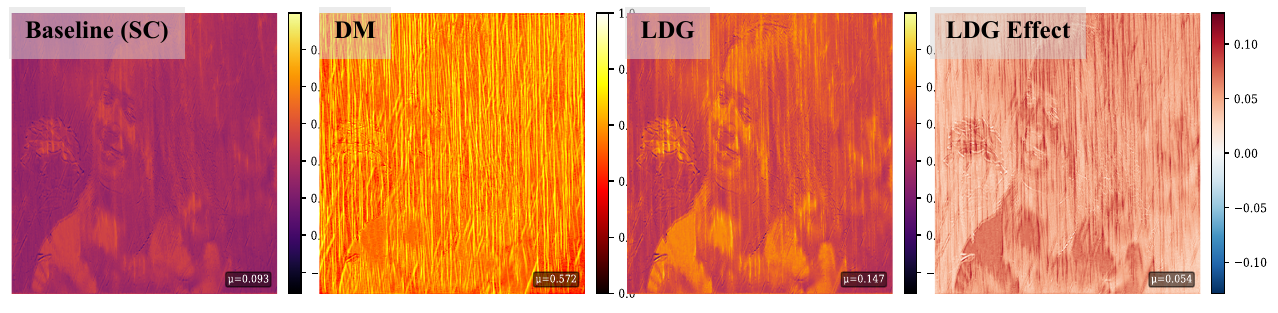}
        \vspace{-.58cm}
        \captionof{figure}{LDG attention visualization.}
        \label{fig:LDG_att}
    \end{minipage}
    \vspace{-0.7cm}
    
\end{figure}

Tab.~\ref{tab:degmap_abl} illustrates the effectiveness of our proposed LDG, which significantly improves restoration performance by localizing the underlying degradation and fusing it with decoder features. Notably, LDG outperforms the baseline skip connection (SC) by 1.30 dB. Fig.~\ref{fig:LDG_att} visually confirms that LDG explicitly isolates and highlights the degradation patterns (e.g., rain streaks) that the baseline skip connection fails to capture.  More analysis is provided in the \textbf{Supp. S4.4}.

\subsection{Real-world Implications}
\begin{table}[!htb]
\centering
\begin{minipage}{0.57\textwidth}
    \centering
\centering
\caption{OD in Lowlight environment.}
\label{tab:od}


\scalebox{0.71}{%
\begin{tabular}{lcccccc}
\toprule
\multirow{2}{*}{\textbf{Method}} & \multirow{2}{*}{\textbf{mAP@50}} & \multicolumn{3}{c}{\textbf{Inference Speed}} & \multirow{2}{*}{\textbf{Overhead}} & \textbf{Efficacy} \\
\cmidrule(lr){3-5}
 & & IR & OD & Total & & \textbf{Score} \\
\midrule

ExDark & 0.256 & - & \multirow{4}{*}{60} & 60 & 0 & - \\
PTQ & 0.203 & 442 & & 53 & 7 & 10.76 \\
FP32 (Ours) & 0.410 & 136 & & 42 & 18 & 17.22 \\

\rowcolor{rowgray}
\textbf{QDR (Ours)} & 
\begin{tabular}[c]{@{}c@{}}\best{0.366}\\{\scriptsize\textcolor{green!60!black}{+16.3\%}}\end{tabular} & 
442 & & 53 & 7 & \best{19.40} \\

\bottomrule
\end{tabular}
}

\end{minipage}
\hfill
\begin{minipage}{0.4\textwidth}
    \centering
\centering
\caption{On-device sustain performance analysis for EVA.}
\label{tab:jetsoninfer}

\setlength{\tabcolsep}{6pt} 
\renewcommand{\arraystretch}{1.2}

\scalebox{0.73}{%
\begin{tabular}{l c c c c}
\toprule
\textbf{Prec.} & \textbf{Clock} & \textbf{Temp} & \textbf{Latency} & \textbf{FPS} \\
 & \textbf{(MHz)} & \textbf{(\ensuremath{^\circ}C)} & \textbf{(ms)} & \\
\midrule
FP32 & 1808.95 & 67.65 & 7.36 & 136 \\
FP16 & 1786.92 & 66.44 & 4.88 & 205 \\
\rowcolor{rowgray}
\textbf{INT8} & \textbf{1900.65} & \textbf{63.33} & \textbf{2.55} & \textbf{442} \\
\bottomrule
\end{tabular}%
}

\end{minipage}
\vspace{-0.3cm}
\end{table}

\subsubsection{IR for Vision} 

To assess practical utility, we deploy our method on an NVIDIA Jetson to evaluate real-world practicability. We utilize our model as an IR pre-processor for YOLOv5 \cite{jocher2020yolov5} object detection on the low-light ExDark dataset~\cite{loh2019getting}. To quantify the accuracy-speed trade-off, we define an Efficacy Score: $\mathcal{E} = \text{mAP@0.50} \times \text{FPS}_{\text{IR+OD}}$. As shown in Tab.~\ref{tab:od}, naive PTQ severely degrades accuracy (0.203 mAP), while the FP32 model incurs high latency. Conversely, our QDR model achieves the optimal balance: it boosts mAP@0.50 to 0.366 while maintaining a fast 53 FPS on the edge. This yields the highest efficacy score (19.40), demonstrating DFD as a highly practical module for real-time EVA.

\subsubsection{Edge Performance}

We evaluate real-world edge deployment by benchmarking sustained performance on an NVIDIA Jetson Orin (Tab.~\ref{tab:jetsoninfer}). Our QDR-optimized model significantly outperforms its FP32 and FP16 counterparts, achieving 442 FPS with a 2.55 ms latency. Notably, it sustains the highest clock speed (1900.65 MHz) at the lowest operating temperature (63.33 $^\circ$C), highlighting its exceptional thermal and computational efficiency for real-time EVA (see \textbf{Supp S2}).

\section{Conclusion}
In this paper, we introduce the QDR framework for real-world EVA, serving as a foundational baseline for QAT-KD in IR. To successfully deploy quantized IR models, we systematically addressed three critical design questions. First, for teacher selection, we demonstrated the efficacy of self-distillation using an FP32 counterpart. Second, to determine the optimal distillation location, we proposed DFD to prevent spatial error amplification. Third, to stabilize this process, we introduced dynamic magnitude balancing, effectively resolving the optimization "tug-of-war" between distillation and task-specific restoration losses in QAT settings. Additionally, we proposed EFM with LDG to dynamically modulate spatial features based on localized degradation, maximizing hardware efficiency. We conducted extensive evaluations on real-world edge platforms and demonstrated that QDR delivers significant performance gains in EVA (e.g., +16.3 mAP on OD) with minimal overhead.  In future work, we plan to extend this framework to all-in-one IR, compound degradation, and diverse hardware architectures. For more results and analysis, please see the supplementary material.


%

\bibliographystyle{splncs04}

\begin{thebibliography}{10}
\providecommand{\url}[1]{\texttt{#1}}
\providecommand{\urlprefix}{URL }
\providecommand{\doi}[1]{https://doi.org/#1}

\bibitem{a2024learning}
A~Sharif, S., Myrzabekov, A., Khudjaev, N., Tsoy, R., Kim, S., Lee, J.: Learning optimized low-light image enhancement for edge vision tasks. In: Proceedings of the IEEE/CVF Conference on Computer Vision and Pattern Recognition. pp. 6373--6383 (2024)

\bibitem{abdelhamed2018high}
Abdelhamed, A., Lin, S., Brown, M.S.: A high-quality denoising dataset for smartphone cameras. In: Proceedings of the IEEE conference on computer vision and pattern recognition. pp. 1692--1700 (2018)

\bibitem{ayazoglu2021extremely}
Ayazoglu, M.: Extremely lightweight quantization robust real-time single-image super resolution for mobile devices. In: Proceedings of the IEEE/CVF conference on computer vision and pattern recognition. pp. 2472--2479 (2021)

\bibitem{boo2021stochastic}
Boo, Y., Shin, S., Choi, J., Sung, W.: Stochastic precision ensemble: self-knowledge distillation for quantized deep neural networks. In: Proceedings of the AAAI Conference on Artificial Intelligence. vol.~35, pp. 6794--6802 (2021)

\bibitem{cai2023retinexformer}
Cai, Y., Bian, H., Lin, J., Wang, H., Timofte, R., Zhang, Y.: Retinexformer: One-stage retinex-based transformer for low-light image enhancement. In: Proceedings of the IEEE/CVF international conference on computer vision. pp. 12504--12513 (2023)

\bibitem{cohen2025efficient}
Cohen, E., Achituve, I., Diamant, I., Netzer, A., Habi, H.V.: Efficient image restoration via latent consistency flow matching. arXiv preprint arXiv:2502.03500  (2025)

\bibitem{conde2023perceptual}
Conde, M.V., Vasluianu, F., Vazquez-Corral, J., Timofte, R.: Perceptual image enhancement for smartphone real-time applications. In: Proceedings of the IEEE/CVF Winter Conference on Applications of Computer Vision. pp. 1848--1858 (2023)

\bibitem{onnxruntime}
developers, O.R.: Onnx runtime. \url{https://onnxruntime.ai/} (2021), version: x.y.z

\bibitem{feijoo2025darkir}
Feijoo, D., Benito, J.C., Garcia, A., Conde, M.V.: Darkir: Robust low-light image restoration. In: Proceedings of the Computer Vision and Pattern Recognition Conference. pp. 10879--10889 (2025)

\bibitem{feng2025q}
Feng, W., Yang, C., Qin, H., Li, X., Wang, Y., An, Z., Huang, L., Diao, B., Zhao, Z., Xu, Y., et~al.: Q-vdit: Towards accurate quantization and distillation of video-generation diffusion transformers. arXiv preprint arXiv:2505.22167  (2025)

\bibitem{hinton2015distilling}
Hinton, G., Vinyals, O., Dean, J.: Distilling the knowledge in a neural network. arXiv preprint arXiv:1503.02531  (2015)

\bibitem{hong2022daq}
Hong, C., Kim, H., Baik, S., Oh, J., Lee, K.M.: Daq: Channel-wise distribution-aware quantization for deep image super-resolution networks. In: Proceedings of the IEEE/CVF Winter Conference on Applications of Computer Vision. pp. 2675--2684 (2022)

\bibitem{hu2018squeeze}
Hu, J., Shen, L., Sun, G.: Squeeze-and-excitation networks. In: Proceedings of the IEEE conference on computer vision and pattern recognition. pp. 7132--7141 (2018)

\bibitem{ioffe2015batch}
Ioffe, S., Szegedy, C.: Batch normalization: Accelerating deep network training by reducing internal covariate shift. In: International conference on machine learning. pp. 448--456. pmlr (2015)

\bibitem{jocher2020yolov5}
Jocher, G., et~al.: ultralytics/yolov5: Initial release. \url{https://doi.org/10.5281/zenodo.3908559} (Jun 2020). \doi{10.5281/zenodo.3908559}

\bibitem{johnson2016perceptual}
Johnson, J., Alahi, A., Fei-Fei, L.: Perceptual losses for real-time style transfer and super-resolution. In: European conference on computer vision. pp. 694--711. Springer (2016)

\bibitem{karumbunathan2022nvidia}
Karumbunathan, L.S.: Nvidia jetson agx orin series. A Giant leap forward for robotics and edge AI applications. Technical Brief  (2022)

\bibitem{kim2019qkd}
Kim, E., Choi, S., Kim, S.: Qkd: Knowledge distillation via quantization and denoising. IEEE Transactions on Neural Networks and Learning Systems  \textbf{30}(3),  831--844 (2019)

\bibitem{li2018benchmarking}
Li, B., Ren, W., Fu, D., Tao, D., Feng, D., Zeng, W., Wang, Z.: Benchmarking single-image dehazing and beyond. IEEE transactions on image processing  \textbf{28}(1),  492--505 (2018)

\bibitem{li2020pams}
Li, H., Yan, C., Lin, S., Zheng, X., Zhang, B., Yang, F., Ji, R.: Pams: Quantized super-resolution via parameterized max scale. In: European conference on computer vision. pp. 564--580. Springer (2020)

\bibitem{li2021dynamic}
Li, L., Lin, Y., Ren, S., Li, P., Zhou, J., Sun, X.: Dynamic knowledge distillation for pre-trained language models. In: Proceedings of the 2021 Conference on Empirical Methods in Natural Language Processing. pp. 379--389 (2021)

\bibitem{loh2019getting}
Loh, Y.P., Chan, C.S.: Getting to know low-light images with the exclusively dark dataset. Computer vision and image understanding  \textbf{178},  30--42 (2019)

\bibitem{lu2021rw}
Lu, P., Ghaddar, A., Rashid, A., Rezagholizadeh, M., Ghodsi, A., Langlais, P.: Rw-kd: Sample-wise loss terms re-weighting for knowledge distillation. In: Findings of the Association for Computational Linguistics: EMNLP 2021. pp. 3145--3152 (2021)

\bibitem{makhov2025prefilt}
Makhov, D., Ostapets, R., Zhelavskaya, I., Song, D.: Prefilt: Prefiltering for fully quantized image restoration neural networks. In: Proceedings of the IEEE/CVF International Conference on Computer Vision. pp. 4042--4051 (2025)

\bibitem{nvidia_model_optimizer_2025}
{NVIDIA}: Nvidia model optimizer. \url{https://github.com/NVIDIA/Model-Optimizer} (2025), gitHub repository, accessed 2026-03-09

\bibitem{onnx_github_2026}
{ONNX}: Onnx. \url{https://github.com/onnx/onnx}, gitHub repository, accessed 2026-03-09

\bibitem{paszke2019pytorch}
Paszke, A., Gross, S., Massa, F., Lerer, A., Bradbury, J., Chanan, G., Killeen, T., Lin, Z., Gimelshein, N., Antiga, L., et~al.: Pytorch: An imperative style, high-performance deep learning library. Advances in neural information processing systems  \textbf{32} (2019)

\bibitem{pham2023cmtkd}
Pham, Q.H., et~al.: Cmt-kd: Collaborative mutual teaching for quantized knowledge distillation. In: Proceedings of the IEEE/CVF Conference on Computer Vision and Pattern Recognition (CVPR) (2023)

\bibitem{pierro2024mamba}
Pierro, A., Abreu, S.: Mamba-ptq: Outlier channels in recurrent large language models. arXiv preprint arXiv:2407.12397  (2024)

\bibitem{rehman2025punching}
Rehman, A., Sharif, S., Rahaman, M.A., Rasool, M.J.A., Kim, S., Lee, J.: Punching above precision: Small quantized model distillation with learnable regularizer. arXiv preprint arXiv:2509.20854  (2025)

\bibitem{ronneberger2015u}
Ronneberger, O., Fischer, P., Brox, T.: U-net: Convolutional networks for biomedical image segmentation. In: International Conference on Medical image computing and computer-assisted intervention. pp. 234--241. Springer (2015)

\bibitem{sharif2026illuminating}
Sharif, S., Rehman, A., Abidin, Z.U., Dharejo, F.A., Timofte, R., Naqvi, R.A.: Illuminating darkness: Learning to enhance low-light images in-the-wild. In: Proceedings of the IEEE/CVF Winter Conference on Applications of Computer Vision. pp. 2263--2272 (2026)

\bibitem{wang2021fully}
Wang, H., Chen, P., Zhuang, B., Shen, C.: Fully quantized image super-resolution networks. In: Proceedings of the 29th ACM International Conference on Multimedia. pp. 639--647 (2021)

\bibitem{wang2022uformer}
Wang, Z., Cun, X., Bao, J., Zhou, W., Liu, J., Li, H.: Uformer: A general u-shaped transformer for image restoration. In: Proceedings of the IEEE/CVF conference on computer vision and pattern recognition. pp. 17683--17693 (2022)

\bibitem{wei2018deep}
Wei, C., Wang, W., Yang, W., Liu, J.: Deep retinex decomposition for low-light enhancement. arXiv preprint arXiv:1808.04560  (2018)

\bibitem{wu2024ptq4dit}
Wu, J., Wang, H., Shang, Y., Shah, M., Yan, Y.: Ptq4dit: Post-training quantization for diffusion transformers. Advances in neural information processing systems  \textbf{37},  62732--62755 (2024)

\bibitem{yang2017deep}
Yang, W., Tan, R.T., Feng, J., Liu, J., Guo, Z., Yan, S.: Deep joint rain detection and removal from a single image. In: Proceedings of the IEEE conference on computer vision and pattern recognition. pp. 1357--1366 (2017)

\bibitem{yang2022masked}
Yang, Z., Li, Z., Shao, M., Shi, D., Yuan, Z., Yuan, C.: Masked generative distillation. In: European conference on computer vision. pp. 53--69. Springer (2022)

\bibitem{yuan2024fakd}
Yuan, J., Phan, M.H., Liu, L., Liu, Y.: Fakd: Feature augmented knowledge distillation for semantic segmentation. In: Proceedings of the IEEE/CVF Winter Conference on Applications of Computer Vision. pp. 595--605 (2024)

\bibitem{zamir2022restormer}
Zamir, S.W., Arora, A., Khan, S., Hayat, M., Khan, F.S., Yang, M.H.: Restormer: Efficient transformer for high-resolution image restoration. In: Proceedings of the IEEE/CVF conference on computer vision and pattern recognition. pp. 5728--5739 (2022)

\bibitem{zamir2022learning}
Zamir, S.W., Arora, A., Khan, S., Hayat, M., Khan, F.S., Yang, M.H., Shao, L.: Learning enriched features for fast image restoration and enhancement. IEEE transactions on pattern analysis and machine intelligence  \textbf{45}(2),  1934--1948 (2022)

\bibitem{Zamir2022MIRNetv2}
Zamir, S.W., Arora, A., Khan, S., Hayat, M., Khan, F.S., Yang, M.H., Shao, L.: Learning enriched features for fast image restoration and enhancement. IEEE Transactions on Pattern Analysis and Machine Intelligence (TPAMI)  (2022)

\bibitem{zhang2017beyond}
Zhang, K., Zuo, W., Chen, Y., Meng, D., Zhang, L.: Beyond a gaussian denoiser: Residual learning of deep cnn for image denoising. IEEE transactions on image processing  \textbf{26}(7),  3142--3155 (2017)

\bibitem{zhang2026dlienet}
Zhang, L., Li, Z., Cheng, L., Zhang, Q., Liu, Z., Zhang, X., Xiao, C.: Dlienet: A lightweight low-light image enhancement network via knowledge distillation. Pattern Recognition  \textbf{169},  111777 (2026)

\bibitem{zhang2018unreasonable}
Zhang, R., Isola, P., Efros, A.A., Shechtman, E., Wang, O.: The unreasonable effectiveness of deep features as a perceptual metric. In: Proceedings of the IEEE conference on computer vision and pattern recognition. pp. 586--595 (2018)

\bibitem{zhang2025knowledge}
Zhang, Y., Yan, D.: Knowledge distillation for image restoration: simultaneous learning from degraded and clean images. In: ICASSP 2025-2025 IEEE International Conference on Acoustics, Speech and Signal Processing (ICASSP). pp.~1--5. IEEE (2025)

\bibitem{zhang2025soft}
Zhang, Y., Yan, D.: Soft knowledge distillation with multi-dimensional cross-net attention for image restoration models compression. In: ICASSP 2025-2025 IEEE International Conference on Acoustics, Speech and Signal Processing (ICASSP). pp.~1--5. IEEE (2025)

\bibitem{zhao2024self}
Zhao, K., Zhao, M.: Self-supervised quantization-aware knowledge distillation. arXiv preprint arXiv:2403.11106  (2024)

\bibitem{zhao24d}
Zhao, K., Zhao, M.: Self-supervised quantization-aware knowledge distillation. In: Proceedings of the 27th International Conference on Artificial Intelligence and Statistics. Proceedings of Machine Learning Research, vol.~238, pp. 4375--4383. PMLR (2024), \url{https://proceedings.mlr.press/v238/zhao24d.html}

\bibitem{zheng2021learning}
Zheng, H., Yang, H., Fu, J., Zha, Z.J., Luo, J.: Learning conditional knowledge distillation for degraded-reference image quality assessment. In: Proceedings of the IEEE/CVF International Conference on Computer Vision. pp. 10242--10251 (2021)

\bibitem{zhong2022dynamic}
Zhong, Y., Lin, M., Li, X., Li, K., Shen, Y., Chao, F., Wu, Y., Ji, R.: Dynamic dual trainable bounds for ultra-low precision super-resolution networks. In: European Conference on Computer Vision. pp. 1--18. Springer (2022)

\bibitem{zhou2022lednet}
Zhou, S., Li, C., Change~Loy, C.: Lednet: Joint low-light enhancement and deblurring in the dark. In: European conference on computer vision. pp. 573--589. Springer (2022)

\bibitem{zhou2025dynamic}
Zhou, Y., Qiao, J., Liao, J., Li, W., Li, S., Xie, J., Shen, Y., Hu, J., Lin, S.: Dynamic contrastive knowledge distillation for efficient image restoration. In: Proceedings of the AAAI Conference on Artificial Intelligence. vol.~39, pp. 10861--10869 (2025)

\bibitem{zhu2023quantized}
Zhu, K., He, Y.Y., Wu, J.: Quantized feature distillation for network quantization. In: Proceedings of the AAAI Conference on Artificial Intelligence. vol.~37, pp. 11452--11460 (2023)

\bibitem{zou2024vqcnir}
Zou, W., Gao, H., Ye, T., Chen, L., Yang, W., Huang, S., Chen, H., Chen, S.: Vqcnir: clearer night image restoration with vector-quantized codebook. In: Proceedings of the AAAI conference on artificial intelligence. vol.~38, pp. 7873--7881 (2024)

\end{thebibliography}

\setcounter{section}{0}
\renewcommand{\thesection}{S\arabic{section}}
\renewcommand{\theHsection}{S\arabic{section}} 

\renewcommand{\thetable}{S\arabic{table}}
\renewcommand{\theHtable}{S\arabic{table}}     

\renewcommand{\thefigure}{S\arabic{figure}}
\renewcommand{\theHfigure}{S\arabic{figure}}   

\renewcommand{\theequation}{S\arabic{equation}}
\renewcommand{\theHequation}{S\arabic{equation}} 

\section*{Supplementary Material}
This supplementary material provides comprehensive implementation details and additional experimental analysis to support the findings in the main paper. The content is organized as follows:

\begin{itemize}

    \item \textbf{Section \ref{sec:supp_network}: Network and Training Details.} We elaborate on the architectural design of our EFM, detail the training dynamics, and provide the pseudocode for our LMR (Algorithms 1 \& 2).

    \item \textbf{Section \ref{edge}: Edge Deployment.} We present the practical edge deployment pipeline, covering the ONNX to TensorRT optimization workflow. Additionally, we report hardware stability benchmarks, including temperature and clock speed analysis, on the NVIDIA Jetson Orin platform.

    \item \textbf{Section \ref{visual_results}: Additional Visual Comparisons.} We provide extensive visual results on benchmark datasets under low-bit settings.

    \item \textbf{Section \ref{analysis}: Extended Ablation Studies.} We analyze the impact of hyperparameters and architectural components on restoration performance.

    \item \textbf{Section \ref{limit}: Limitations and Future Work.} We discuss the current limitations and outline promising directions for future research.

\end{itemize}

\section{Network Details}
\label{sec:supp_network}
\subsection{Edge-friendly Model}

\begin{figure}[!htb]
    \centering
    \includegraphics[width=\linewidth]{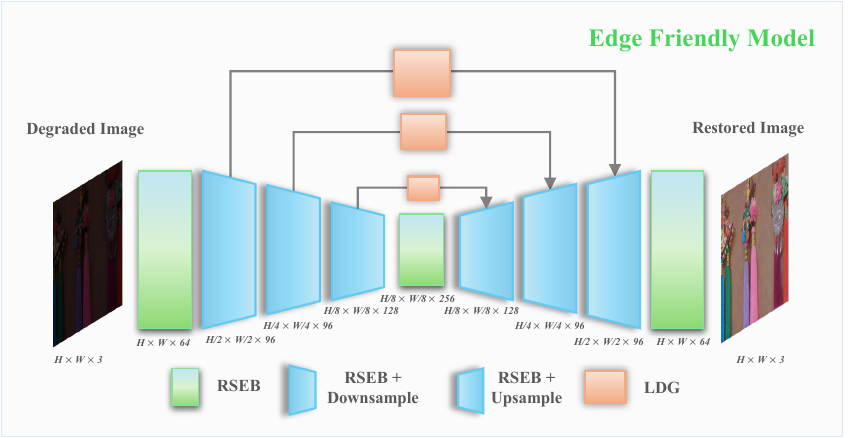}
    \caption{Overview of proposed edge-friendly model.}
    \label{fig:network}
    
\end{figure}

\noindent\textbf{Overview.}
We design a lightweight encoder-bottleneck-decoder network optimized for resource-constrained edge devices. Fig. \ref{fig:network} shows the overview of our proposed EFM. Unlike conventional U-Net\cite{ronneberger2015u} variants, all architectural choices minimize memory access cost (MAC) and maximize quantization compatibility. Therefore, the encoder progressively downsamples via strided convolutions, the bottleneck applies two residual blocks, and the
decoder restores spatial resolution through transposed convolutions
with LDG. Tab.~\ref{tab:arch} summarizes the per-stage dimensions.

\begin{table}[!htb]
\centering
\caption{EFM architecture overview. $H \times W$ denotes the spatial
resolution of the input image.}
\label{tab:arch}
\resizebox{\linewidth}{!}{%
\begin{tabular}{llccc}
\toprule
\textbf{Stage} & \textbf{Module} & \textbf{Channels} &
\textbf{Spatial Size} & \textbf{Operation} \\
\midrule
Input proj.   & Conv $3\times3$              & $3 \to 64$    & $H\times W$                     & Stem convolution \\
\midrule
Encoder 1     & ResBlock + SE                & $64$          & $H\times W$                     & Feature modulation \\
              & MultiScaleDegMap             & $64 \to 1$    & $H\times W$                     & Degradation map $G_1$ \\
              & Conv $3\times3$, stride 2    & $64 \to 96$   & $\frac{H}{2}\times\frac{W}{2}$  & Downsampling \\
\midrule
Encoder 2     & ResBlock + SE                & $96$          & $\frac{H}{2}\times\frac{W}{2}$  & Feature modulation \\
              & MultiScaleDegMap             & $96 \to 1$    & $\frac{H}{2}\times\frac{W}{2}$  & Degradation map $G_2$ \\
              & Conv $3\times3$, stride 2    & $96 \to 128$  & $\frac{H}{4}\times\frac{W}{4}$  & Downsampling \\
\midrule
Encoder 3     & ResBlock + SE                & $128$         & $\frac{H}{4}\times\frac{W}{4}$  & Feature modulation \\
              & MultiScaleDegMap             & $128 \to 1$   & $\frac{H}{4}\times\frac{W}{4}$  & Degradation map $G_3$ \\
              & Conv $3\times3$, stride 2    & $128 \to 256$ & $\frac{H}{8}\times\frac{W}{8}$  & Downsampling \\
\midrule
Bottleneck    & ResBlock + SE $\times 2$     & $256$         & $\frac{H}{8}\times\frac{W}{8}$  & Dense feature refinement \\
\midrule
Decoder 3     & ConvTranspose $2\times2$, stride 2 & $256 \to 128$ & $\frac{H}{4}\times\frac{W}{4}$ & Upsampling \\
              & Scalar fusion ($\alpha_3$)   & $128$         & $\frac{H}{4}\times\frac{W}{4}$  & Skip connection \\
              & ResBlock + SE                & $128$         & $\frac{H}{4}\times\frac{W}{4}$  & Feature modulation \\
\midrule
Decoder 2     & ConvTranspose $2\times2$, stride 2 & $128 \to 96$ & $\frac{H}{2}\times\frac{W}{2}$ & Upsampling \\
              & Scalar fusion ($\alpha_2$)   & $96$          & $\frac{H}{2}\times\frac{W}{2}$  & Skip connection \\
              & ResBlock + SE                & $96$          & $\frac{H}{2}\times\frac{W}{2}$  & Feature modulation \\
\midrule
Decoder 1     & ConvTranspose $2\times2$, stride 2 & $96 \to 64$  & $H\times W$            & Upsampling \\
              & Scalar fusion ($\alpha_1$)   & $64$          & $H\times W$             & Skip connection \\
\midrule
Output proj.  & Conv $1\times1$ + residual   & $64 \to 3$    & $H\times W$             & Global residual add \\
\bottomrule
\end{tabular}}
\end{table}

\noindent\textbf{Residual Block with Squeeze-and-Excitation (ResBlock + SE).}
Each feature modulation block consists of two $3\times3$ convolutions
with \texttt{BatchNorm} \cite{ioffe2015batch} (momentum $0.8$) and \texttt{ReLU}, followed
by an SE channel attention layer~\cite{hu2018squeeze} with reduction
ratio $r{=}16$, and an identity shortcut:
\begin{equation}
    \mathbf{F}(x) = x + \text{SE}\!\left(\text{BN}(\text{Conv}_{3\times3}(
    \text{ReLU}(\text{BN}(\text{Conv}_{3\times3}(x)))))\right)
\end{equation}
\texttt{ReLU} is piecewise-linear and natively fused with convolution
as a single operator in all major NPU toolchains (\emph{e.g.},
TensorRT, TFLite, SNPE), producing activations with a trivially quantizable range $[0, x_{\max}]$ that does not require special calibration handling.

\noindent\textbf{Learnable Degradation Gated Fusion.}
At each encoder stage, a bias-free $1\times1$ convolution projects to $C/8$ channels, followed by a $3\times3$ convolution with Sigmoid, producing a spatial degradation map $G_i \in [0,1]^{H_i \times W_i}$. This LDG map then guides the skip connection fusion:

\begin{equation}
    X_{\text{out}} = X_{\text{up}} + \alpha_i \cdot X_{\text{enc}} \cdot (\alpha_i + G_i)
\end{equation}

where $\alpha_i$ is a learnable scalar. We replaced the standard concatenation-plus-$1\times1$-convolution pattern with element-wise operations, eliminating the $2C \times H \times W$ buffer allocation. It  reduced extra memory from $\mathcal{O}(CHW)$ to $\mathcal{O}(HW)$ per stage. The hard-bounded output $G_i \in [0,1]$ fixes the INT8 scale factor at $s {=} 1/127$ with no per-tensor calibration, and bias-free convolutions eliminate the zero-point correction pass required by most edge NPUs.

\noindent\textbf{Quantization-Friendly Output Layer.}
To avoid INT8 overflow artifacts and costly edge ALU approximations, commonly observed with \texttt{Tanh}, we adopts a standard linear global residual path ($Y_{\text{pred}} = \text{Conv}_{1\times1}(\text{ReLU}(X_{\text{final}})) + X_{\text{in}}$).

\subsection{Quantization}
We adopt a standard INT8 quantization configuration as our baseline. Weights are quantized using a per-channel scheme to preserve fine-grained numerical fidelity, while activations utilize per-tensor quantization with a single global scale factor. Calibration ranges are determined via the max calibration rule~\cite{nvidia_model_optimizer_2025}. We emphasize that our work focuses on developing QDR rather than introducing novel quantization schemes; therefore, we leverage the established configuration to target efficient edge deployment. INT8 serves as our primary precision due to hardware constraints; however, we further evaluate INT4 and INT2 configurations to systematically characterize the trade-off between model compactness and restoration quality.

\subsection{Training Details}
The reconstruction objective is defined as the $\ell_1$ loss between the restored image and the ground-truth target. When knowledge distillation is enabled, we use a fixed pretrained teacher and impose an additional feature distillation loss, implemented as the mean-squared error between the final-layer teacher and student features. The overall training objective is obtained by combining the reconstruction and distillation terms through our LMR strategy. Algo. \ref{alg:init} and Algo. \ref{alg:lmr} details the LMR initialization and training strategy leveraged while learning QDR.


\subsubsection{Algorithm}
\label{sec:lmr}

\begin{algorithm}[!htb]
\DontPrintSemicolon
\caption{LMR: Gradient-Norm Initialization of Learnable Regularizers}
\label{alg:init}
\KwIn{student $f_s(\cdot;\theta_S)$, teacher $f_t(\cdot;\phi)$, data loader $\mathcal{D}$, calibration batches $N$, momentum $\mu$}
\KwOut{learnable log-space parameters $\alpha, \beta \in \mathbb{R}$, EMA seeds $\bar{g}_{\mathrm{rec}}^{(0)}, \bar{g}_{\mathrm{kd}}^{(0)}$}

$\mathcal{G}_{\mathrm{rec}} \leftarrow [\,]$,\quad $\mathcal{G}_{\mathrm{kd}} \leftarrow [\,]$ \;
\ForEach{mini-batch $\{(x_i, y_i)\} \subset \mathcal{D}$, up to $N$ batches}{
    $\hat{y}_s,\, F_s \leftarrow f_s(x;\theta_S)$;\qquad $\hat{y}_t,\, F_t \leftarrow f_t(x;\phi)$ \tcp*{teacher: no grad}
    $\mathcal{L}_{\mathrm{QR}} \leftarrow \|\hat{y}_s - y\|_1$;\qquad $g_{\mathrm{rec}} \leftarrow \|\nabla_{\theta_S}\,\mathcal{L}_{\mathrm{QR}}\|_2$;\qquad Append $g_{\mathrm{rec}}$ to $\mathcal{G}_{\mathrm{rec}}$ \;
    $\mathcal{L}_{\mathrm{KD}} \leftarrow \|F_s - F_t\|_2^2$;\qquad $g_{\mathrm{kd}} \leftarrow \|\nabla_{\theta_S}\,\mathcal{L}_{\mathrm{KD}}\|_2$;\qquad Append $g_{\mathrm{kd}}$ to $\mathcal{G}_{\mathrm{kd}}$ \;
}

\BlankLine
\tcp*{Compute mean norms and apply inverse weighting}
\BlankLine

$\bar{g}_{\mathrm{rec}} \leftarrow \mathrm{mean}(\mathcal{G}_{\mathrm{rec}})$,\quad $\bar{g}_{\mathrm{kd}} \leftarrow \mathrm{mean}(\mathcal{G}_{\mathrm{kd}})$,\quad $\gamma \leftarrow \bar{g}_{\mathrm{rec}} + \bar{g}_{\mathrm{kd}}$ \tcp*{normalization}
$\alpha \leftarrow \log\!\left(\bar{g}_{\mathrm{kd}} / \gamma\right)$,\qquad $\beta \leftarrow \log\!\left(\bar{g}_{\mathrm{rec}} / \gamma\right)$ \tcp*{$\lambda_{\mathrm{rec}}{=}e^\alpha$, $\lambda_{\mathrm{kd}}{=}e^\beta$, jointly optimized with $\theta_S$}
$\bar{g}_{\mathrm{rec}}^{(0)} \leftarrow \bar{g}_{\mathrm{rec}}$,\quad $\bar{g}_{\mathrm{kd}}^{(0)} \leftarrow \bar{g}_{\mathrm{kd}}$ \tcp*{seed EMA buffers, gives $s(0)\approx 1$}

\BlankLine
\tcp{$\alpha$ and $\beta$ are registered as \textbf{nn.Parameter} and added to the optimizer}
\end{algorithm}

\begin{algorithm}[!htb]
\DontPrintSemicolon
\caption{LMR: Training with EMA Gradient-Balanced KD}
\label{alg:lmr}
\KwIn{student $f_s(\cdot;\theta_S)$, frozen teacher $f_t(\cdot;\phi)$, data loader $\mathcal{D}$, optimizer $\mathcal{O}$, total steps $T$, EMA refresh interval $T_g$, momentum $\mu$, grad-clip $c$, learning rates $\eta_{\theta}, \eta_r$}
\KwOut{optimized student $\theta_S^\star$, learned regularizers $\alpha^\star, \beta^\star$}

\tcp{Initialize via Algorithm~\ref{alg:init}}
$\alpha,\, \beta,\, \bar{g}_{\mathrm{rec}}^{(0)},\, \bar{g}_{\mathrm{kd}}^{(0)} \leftarrow \textbf{Algorithm~\ref{alg:init}}(\theta_S, \phi, \mathcal{D})$;\quad Add $\{\alpha, \beta\}$ to $\mathcal{O}$ with lr $\eta_r$ \;
\BlankLine
\ForEach{step $t = 1$ \KwTo $T$}{
    $(x, y) \sim \mathcal{D}$;\qquad $\hat{y}_s,\, F_s \leftarrow f_s(x;\theta_S)$;\qquad $\hat{y}_t,\, F_t \leftarrow f_t(x;\phi)$ \tcp*{teacher: no grad}
    $\mathcal{L}_{\mathrm{QR}} \leftarrow \|\hat{y}_s - y\|_1$;\qquad $\mathcal{L}_{\mathrm{KD}} \leftarrow \|F_s - F_t\|_2^2$ \;
    \BlankLine
    \tcp*{Refresh EMA gradient norms every $T_g$ steps, reuse otherwise}
    \eIf{$t \bmod T_g = 0$}{
        $g_{\mathrm{rec}}(t) \leftarrow \|\nabla_{\theta_S}\,\mathcal{L}_{\mathrm{QR}}\|_2$;\qquad $g_{\mathrm{kd}}(t) \leftarrow \|\nabla_{\theta_S}\,\mathcal{L}_{\mathrm{KD}}\|_2$ \;
        $\bar{g}_{\mathrm{rec}}(t) \leftarrow \mu\,\bar{g}_{\mathrm{rec}}(t\!-\!1) + (1-\mu)\,g_{\mathrm{rec}}(t)$;\qquad $\bar{g}_{\mathrm{kd}}(t) \leftarrow \mu\,\bar{g}_{\mathrm{kd}}(t\!-\!1) + (1-\mu)\,g_{\mathrm{kd}}(t)$ \;
    }{
        $\bar{g}_{\mathrm{rec}}(t) \leftarrow \bar{g}_{\mathrm{rec}}(t\!-\!1)$;\qquad $\bar{g}_{\mathrm{kd}}(t) \leftarrow \bar{g}_{\mathrm{kd}}(t\!-\!1)$ \tcp*{zero overhead}
    }
    \BlankLine
    \tcp*{Compute smoothed gradient ratio and LMR loss}
    $\lambda_{\mathrm{rec}} \leftarrow e^{\alpha}$,\quad $\lambda_{\mathrm{kd}} \leftarrow e^{\beta}$ \tcp*{log-space ensures $\lambda_{\mathrm{rec}}, \lambda_{\mathrm{kd}} > 0$, no sign flips}
    $s(t) \leftarrow \sqrt{\bar{g}_{\mathrm{kd}}(t)\,/\,(\bar{g}_{\mathrm{rec}}(t) + \epsilon)}$;\qquad $r_t \leftarrow \lambda_{\mathrm{rec}}\,/\,\lambda_{\mathrm{kd}}$ \;
    $\mathcal{L}_{\mathrm{QDR}}^{\mathrm{LMR}} \leftarrow \dfrac{r_t}{s(t)}\,\mathcal{L}_{\mathrm{QR}} + \dfrac{s(t)}{r_t}\,\mathcal{L}_{\mathrm{KD}}$ \;
    \BlankLine
    \tcp*{Joint update of $\theta_S$, $\alpha$, $\beta$ via a single backward pass}
    $\{\theta_S,\, \alpha,\, \beta\} \leftarrow \mathcal{O}.\mathrm{step}\!\left(\nabla\,\mathcal{L}_{\mathrm{QDR}}^{\mathrm{LMR}}\right)$ \;
    $\mathrm{ClipGradNorm}(\{\alpha, \beta\},\, c)$;\qquad $\alpha \leftarrow \max(\alpha,\, \log 10^{-4})$;\qquad $\beta \leftarrow \max(\beta,\, \log 10^{-4})$ \tcp*{clip}
}
\end{algorithm}

\paragraph{Initialization strategy.}
Algorithm~\ref{alg:init} estimates the mean gradient norms of
$\mathcal{L}_{\mathrm{ce}}$ and $\mathcal{L}_{\mathrm{kd}}$ over $N$
calibration batches and applies \emph{inverse weighting}: the regularizer for
each loss is initialized proportionally to the gradient norm of the
\emph{other} loss, so that the two terms enter training with comparable
effective magnitudes.
The EMA buffers are seeded with the same calibration averages, giving
$s^{(0)} \approx 1$ and avoiding a cold-start imbalance.

\paragraph{Complexity.}
Full gradient computation for EMA updates is performed only every $T_g$ steps
(we use $T_g = 50$), incurring an overhead of two additional backward passes
at those steps. Between updates, the stored EMA values are used at zero extra
cost, making the scheme computationally efficient.

\section{Edge Deployment}
\label{edge}

\subsection{On-Device Implementation}
After obtaining the optimized quantized model, we export it to ONNX \cite{onnx_github_2026, onnxruntime} using the PyTorch \cite{paszke2019pytorch} ONNX exporter and convert it into a TensorRT engine with NVIDIA TensorRT. This conversion enables inference-oriented optimizations, including layer fusion, for improved runtime efficiency. We then deploy the engine on an NVIDIA Jetso Orin and report end-to-end latency, memory consumption, and throughput to assess its practicality under real-time edge-device constraints. For a comprehensive hardware evaluation, we optimized EFM alongside SOTA models using standardized APIs to compare their TensorRT FP32 performance under a fair protocol. However, several baseline models (e.g., DarkNet\cite{feijoo2025darkir}, Uformer\cite{wang2022uformer}, ELIR\cite{cohen2025efficient}) contain operations that lack native INT8 support on NVIDIA Orin. Deploying these networks would require non-trivial layer modifications that alter their original design; we omit their INT8 deployment to preserve their architectural integrity and ensure a rigorous comparison.

\subsection{Performance Sustainability}
\begin{figure*}[t]
    \centering
    \begin{subfigure}[b]{0.48\textwidth}
        \centering
        \includegraphics[width=\textwidth]{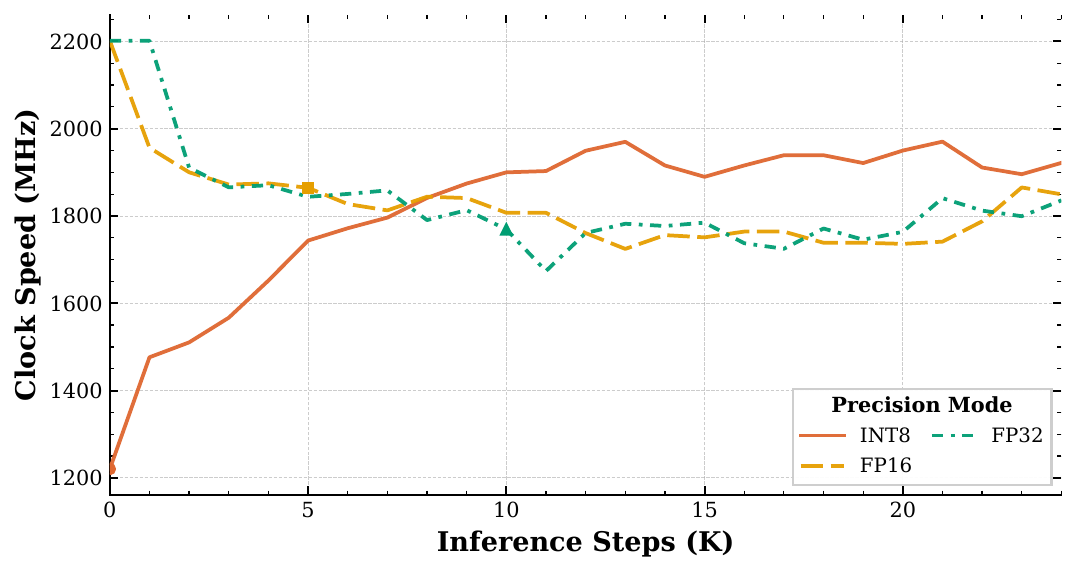}
        \caption{Clock speed over inference steps.}
        \label{fig:temperature}
    \end{subfigure}
    \hfill
    \begin{subfigure}[b]{0.48\textwidth}
        \centering
        \includegraphics[width=\textwidth]{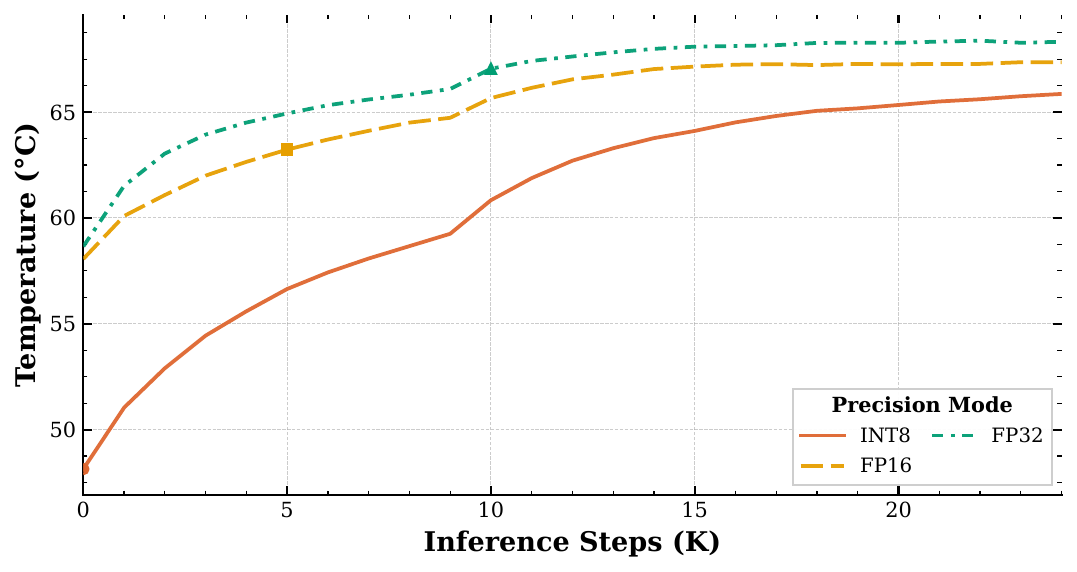}
        \caption{Temperature over inference steps.}
        \label{fig:clockspeed}
    \end{subfigure}
    \caption{Hardware stability during inference.  (a) clock speed and (b) Temperature across inference steps for different precision modes.}
    \label{fig:hardware_metrics}
\end{figure*}

Fig. \ref{fig:hardware_metrics} illustrates the sustained performance of our optimized model compared to standard FP models on the Jetson Orin platform. During prolonged inference, the baseline FP16 and FP32 models rapidly reach thermal limits, triggering frequency throttling. In contrast, our QDR-optimized model maintains a significantly lower operating temperature, allowing the device to sustain higher clock speeds and avoid severe throttling over extended periods.

\subsection{ Downstream Vision Task}
For downstream evaluation, we use the ExDark benchmark \cite{loh2019getting}, which contains 1,473 test images captured under low-light conditions. Each image is annotated with object bounding boxes spanning 12 object categories. To assess whether image restoration improves high-level visual understanding, we first apply each enhancement/restoration method to the ExDark \cite{loh2019getting} test set and then feed the processed images into the object detector. Following prior low-light detection protocols, detection performance is evaluated on the enhanced ExDark \cite{loh2019getting} test images using mean Average Precision (mAP). This setting allows us to directly measure how different restoration methods affect downstream object detection performance under challenging low-light conditions.

\section{More Results}
\label{visual_results}

\begin{figure}[!htb]
    \centering
    \includegraphics[width=\linewidth]{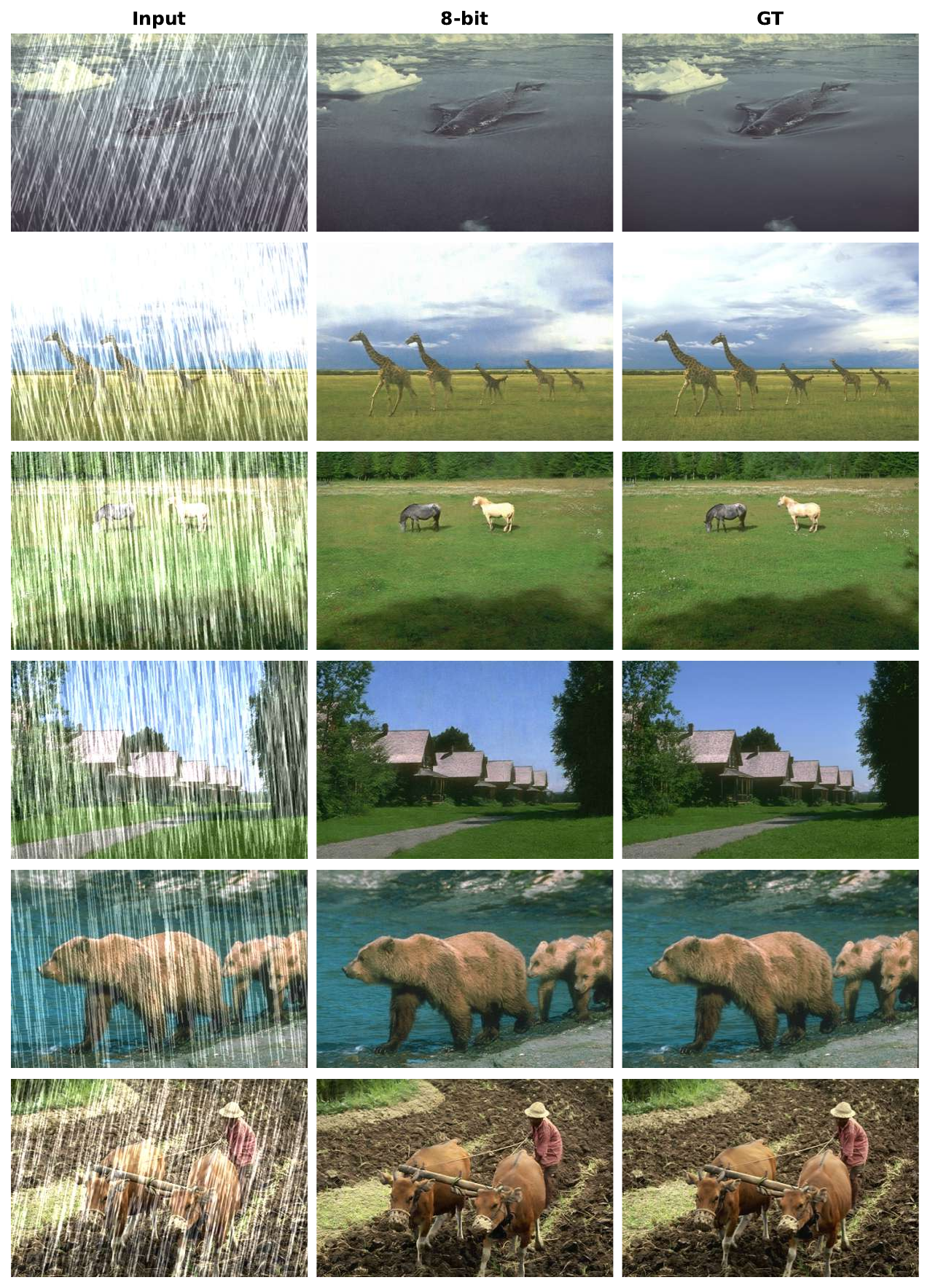}
    \caption{Qualitative comparison of 8-bit image deraining on the Rain100H dataset.}
    \label{fig:rain}
\end{figure}

\begin{figure}[!htb]
    \centering
    \includegraphics[width=\linewidth]{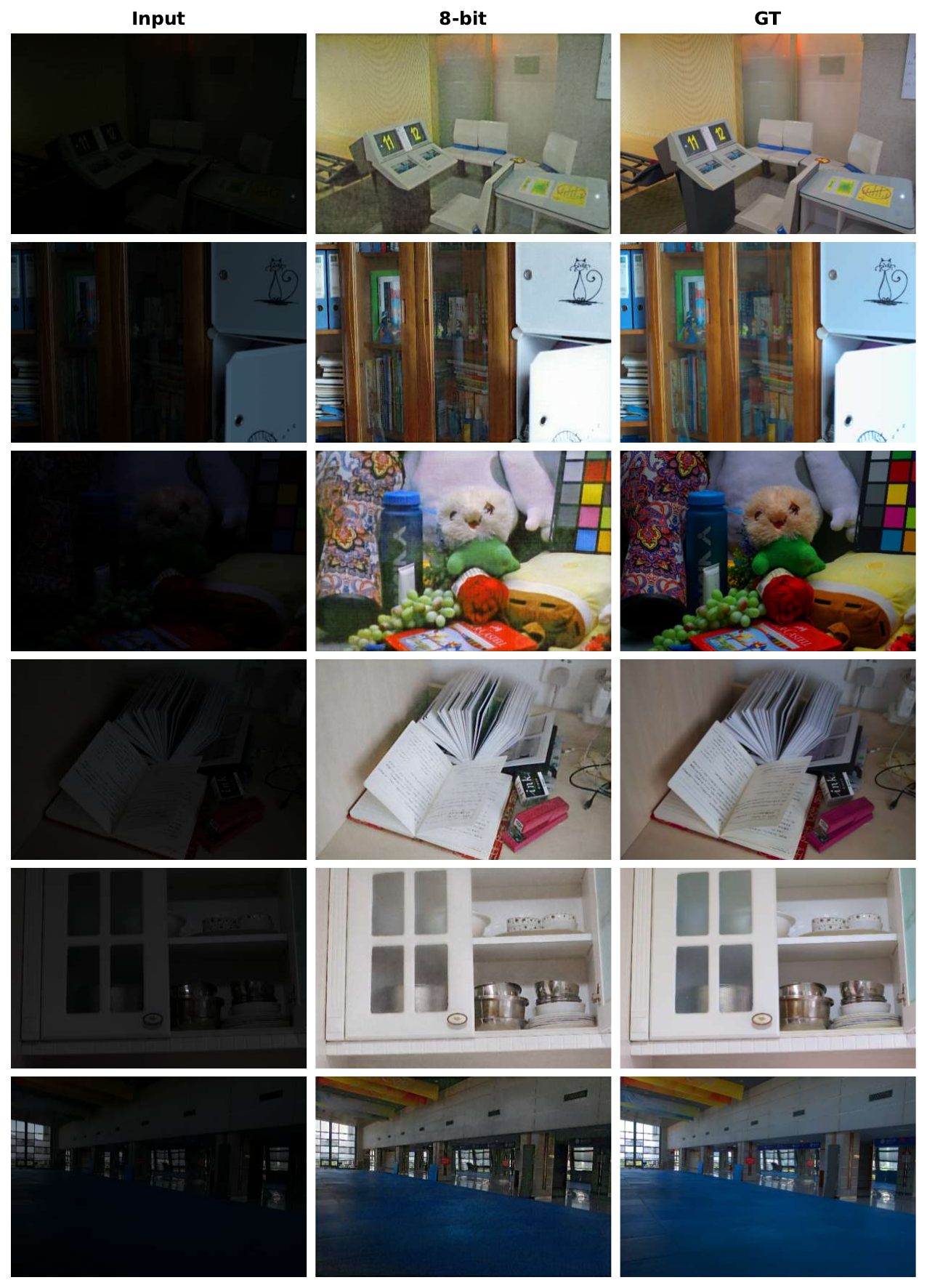}
    \caption{Qualitative comparison of 8-bit low-light enhancement on the LOL dataset.}
    \label{fig:lol}
\end{figure}

\begin{figure}[!htb]
    \centering
    \includegraphics[width=\linewidth]{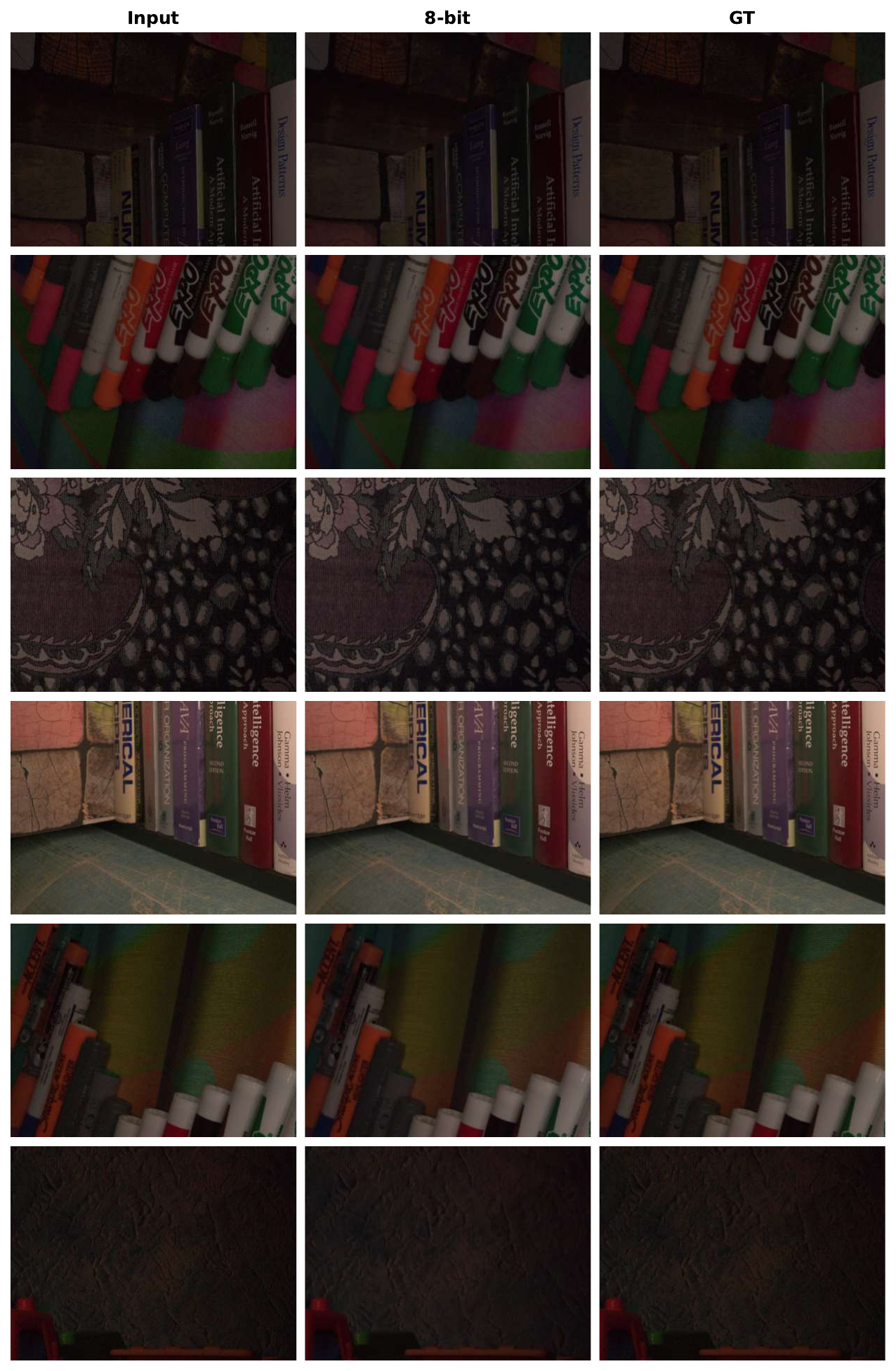}
    \caption{Qualitative comparison of 8-bit real-world denoising on the SIDD dataset.}
    \label{fig:noise}
\end{figure}

\begin{figure}[!htb]
    \centering
    \includegraphics[width=\linewidth]{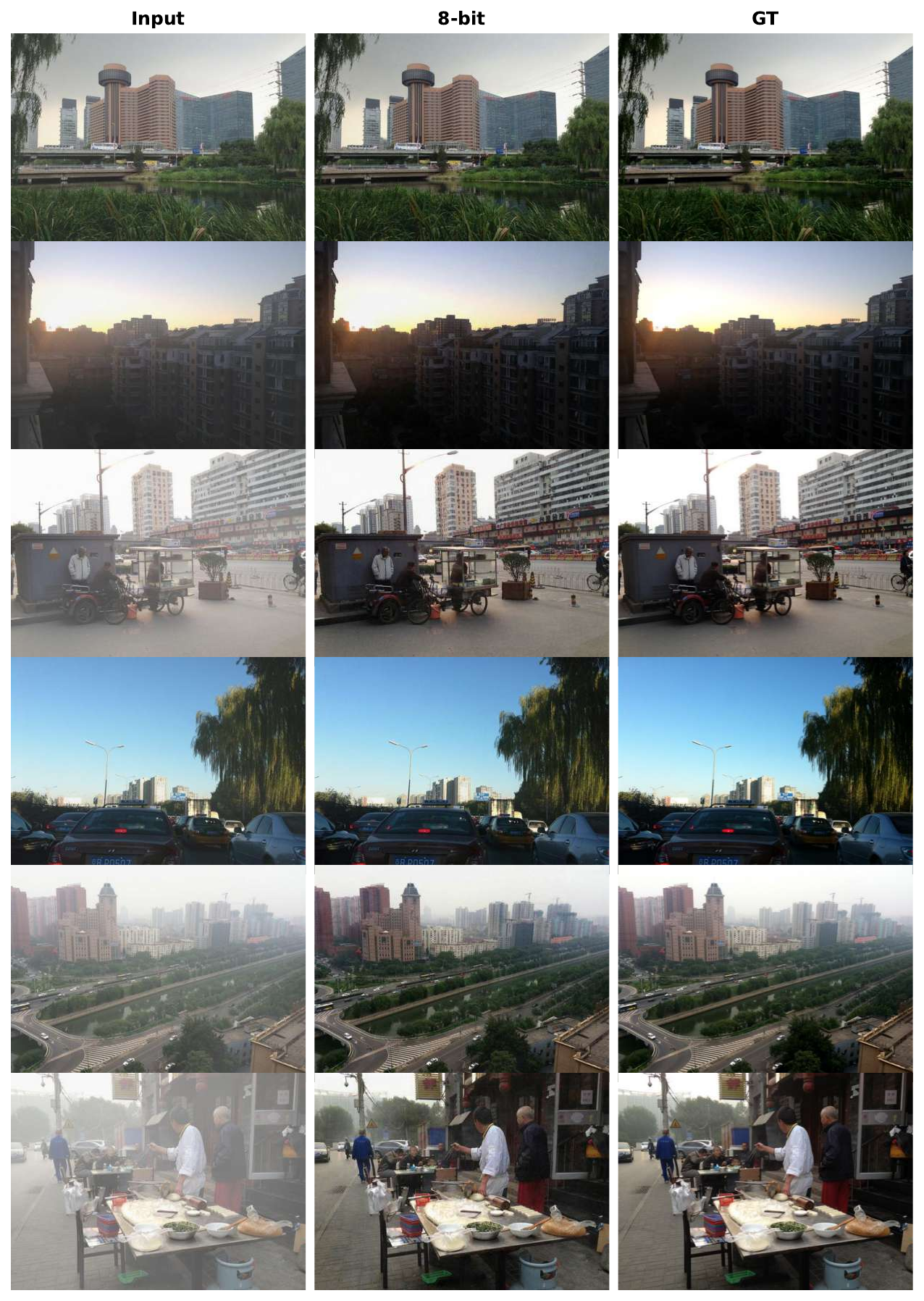}
    \caption{Qualitative comparison of 8-bit image dehazing on the SOTS dataset.}
    \label{fig:haze} 
\end{figure}

We provide additional qualitative comparisons for various restoration tasks in Figs.~\ref{fig:rain}-\ref{fig:haze}. We show 8-bit inference results for image deraining on Rain100H \cite{yang2017deep} (Fig.~\ref{fig:rain}), low-light enhancement on LOL \cite{wei2018deep} (Fig.~\ref{fig:lol}), real-world denoising on SIDD \cite{abdelhamed2018high} (Fig.~\ref{fig:noise}), and image dehazing on SOTS \cite{li2018benchmarking} (Fig.~\ref{fig:haze}).

\section{Analysis}
\label{analysis}

\subsection{PTQ vs QDR}

\begin{figure}[!htb]
    \centering
    \includegraphics[width=\linewidth]{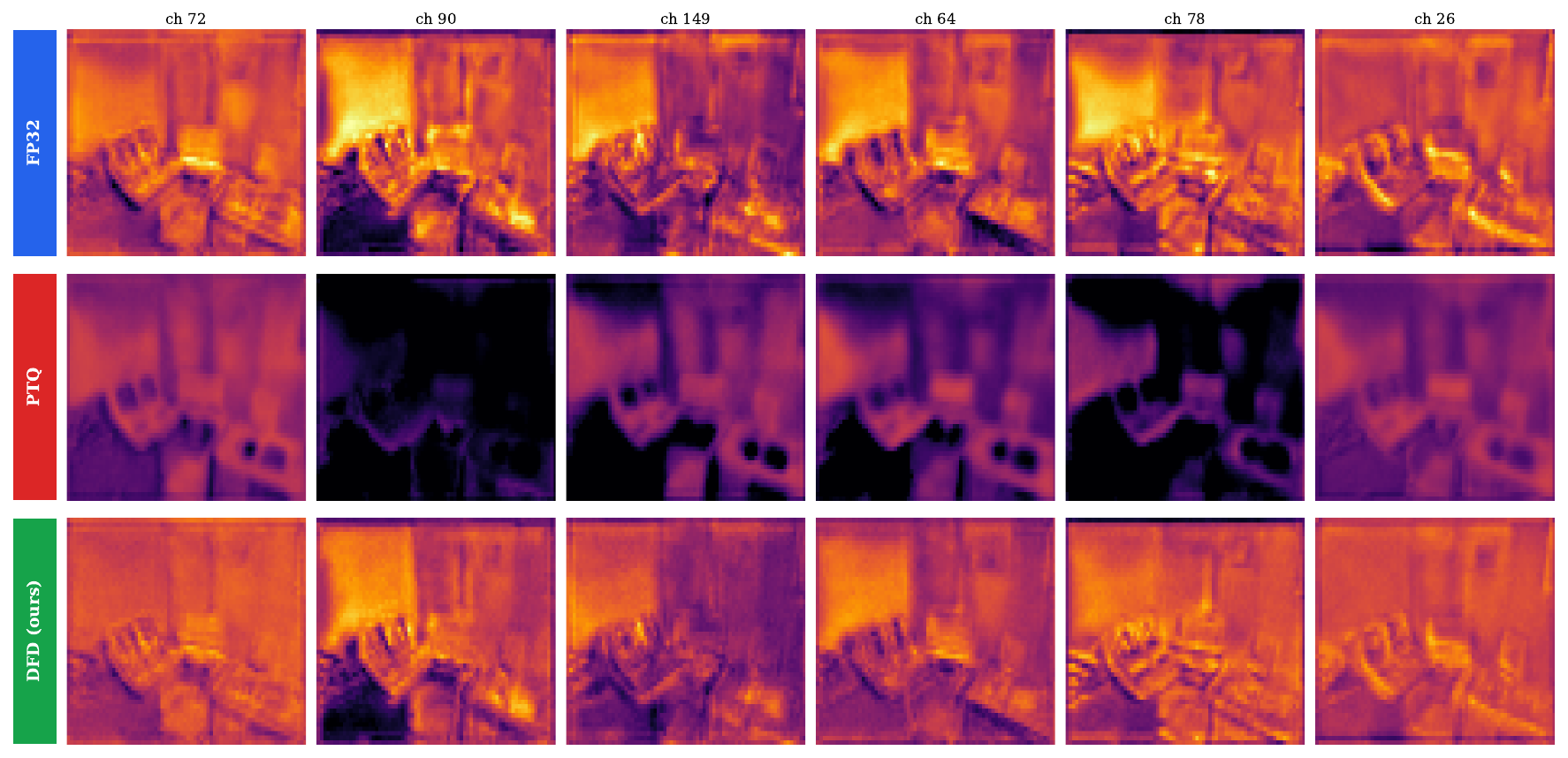}
    \caption{Comparison of activation distributions between PTQ and our QDR method in the bottleneck layers.}
    \label{fig:bottleneck}
\end{figure}

\begin{figure}[!htb]
    \centering
    \includegraphics[width=\linewidth]{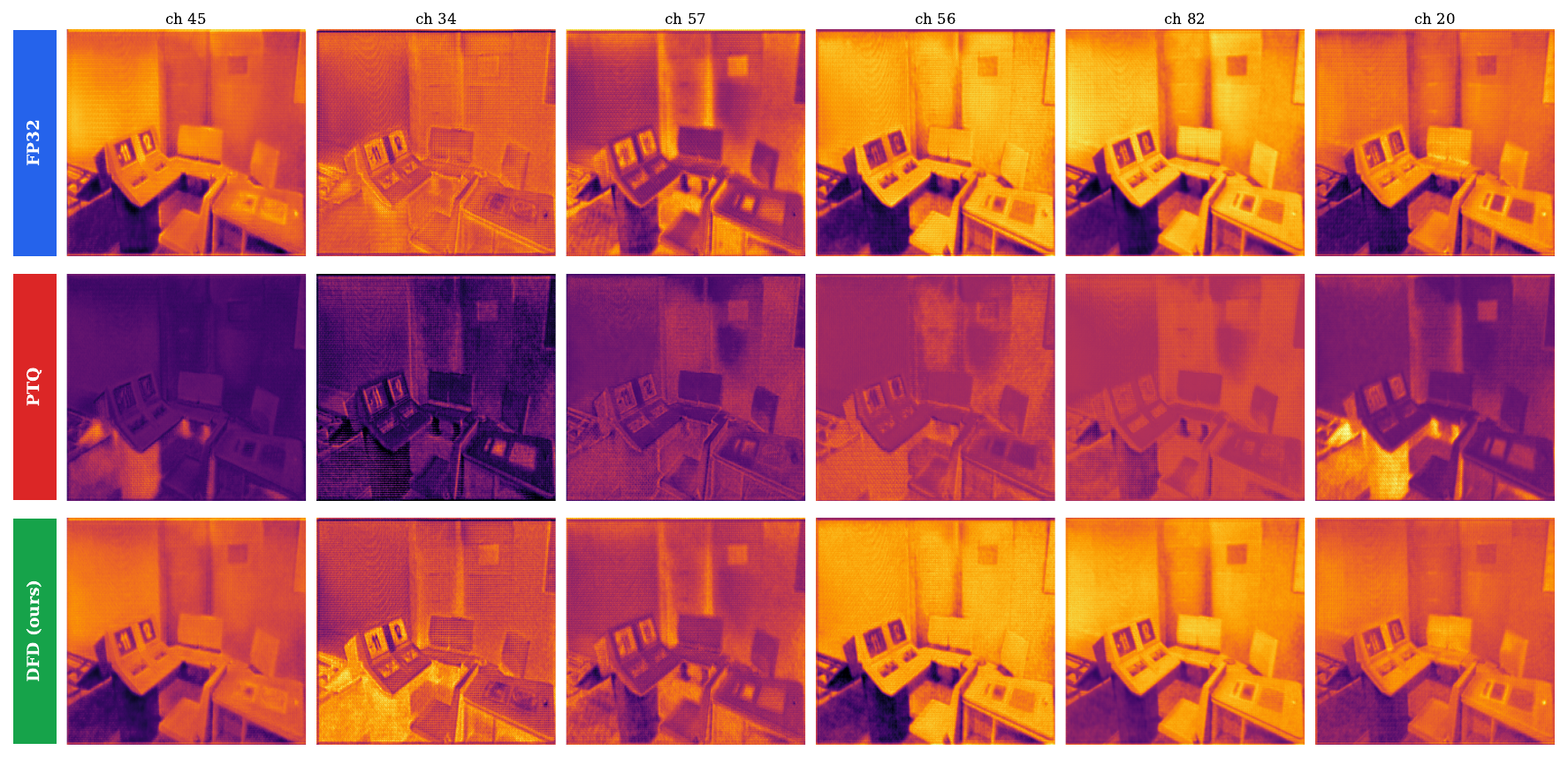}
    \caption{Comparison of activation distributions between PTQ and our QDR method in the decoder layers.}
    \label{fig:decoder}
\end{figure}

To better understand the impact of our method on internal feature representations, we visualize the activation statistics in Figs.~\ref{fig:bottleneck} and \ref{fig:decoder}. Fig.~\ref{fig:bottleneck} illustrates the activation channels within the bottleneck, while Fig.~\ref{fig:decoder} focuses on the decoder layers. As shown, standard PTQ often results in significant distributional deviations, whereas our QDR optimization effectively aligns the distribution with the full-precision model, leading to better perceptual restoration.

\subsection{Teacher Analysis}

\begin{figure}[!htb]
    \centering
    \includegraphics[width=\linewidth]{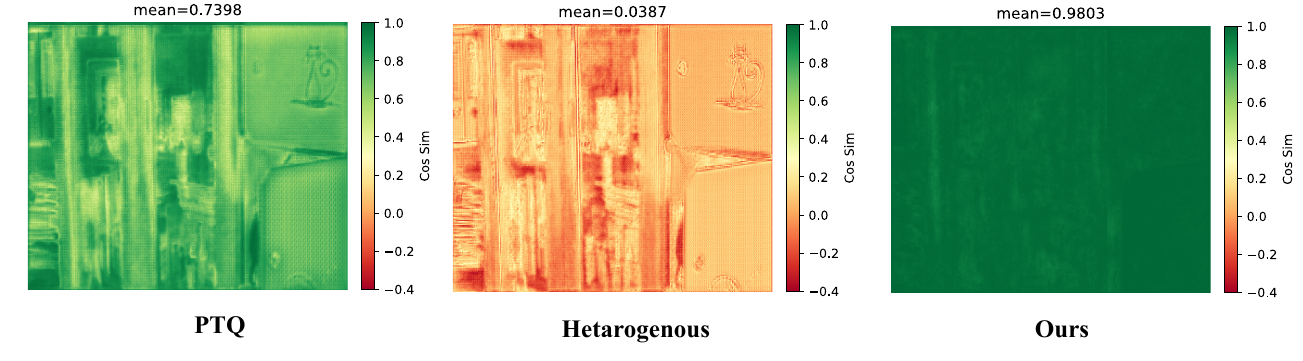}
    \caption{Feature deviation comparison relative to the target FP model. We compare the standard PTQ model, the homogeneous (FP32) teacher-guided model, and the heterogeneous teacher-guided model.}
    \label{fig:teacher}
\end{figure}

Fig.~\ref{fig:teacher} visualizes the feature differences between the quantized models and the target FP32 baseline. While PTQ and homogeneous (FP32) teacher guidance show noticeable feature divergence, our heterogeneous teacher approach significantly reduces this gap, aligning the quantized features more closely with the target representation.

\subsection{Degradation Map Visualization}

\begin{figure}[!htb]
    \centering
    \includegraphics[width=\linewidth]{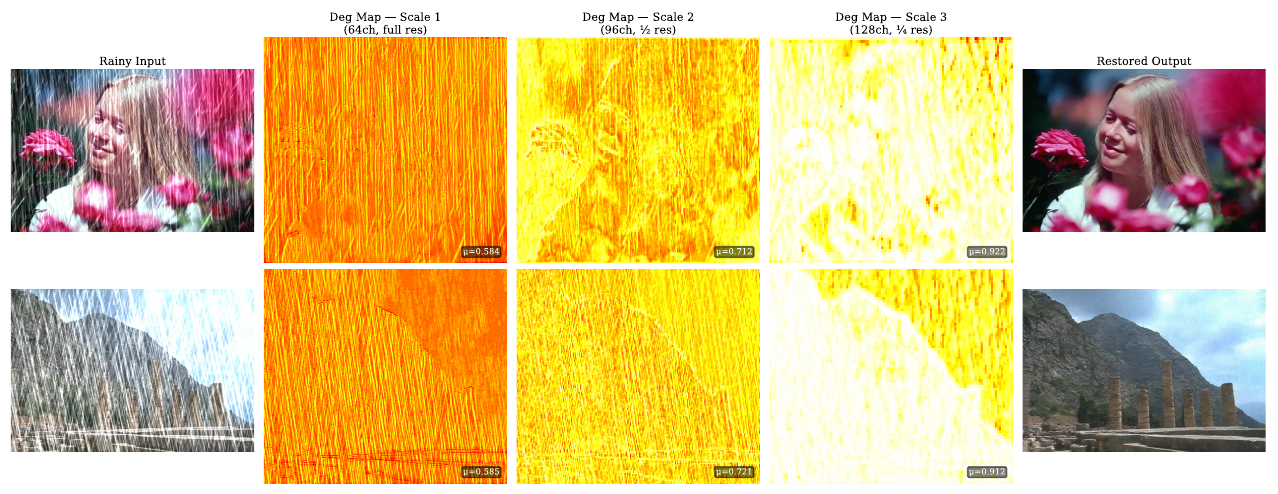}
    \caption{Visualization of degradation maps across different feature scales.}
    \label{fig:degmap}
\end{figure}

We provide a multi-scale visualization of the estimated degradation maps in Fig.~\ref{fig:degmap}. These visualizations demonstrate our model's ability to accurately localize and characterize degradation patterns (e.g., rain streaks) at various feature resolutions, ensuring robust restoration even under low-bit quantization.

\subsection{KD Loss}

To validate the effectiveness of our distillation loss function, we conduct an ablation study comparing it against standard distillation objectives. These include the SOTA KD method (MGD \cite{yang2022masked}), Perceptual Loss \cite{johnson2016perceptual}, and the standard L1 distance. As shown in Table~\ref{tab:loss_ablation}, our simple MSE-based approach significantly outperforms these strategies, achieving a gain of 0.37 dB over the strong L1 baseline and over 2 dB compared to MGD\cite{yang2022masked} and Perceptual Loss\cite{johnson2016perceptual}.

\begin{table}[!htb]
    \centering
    \caption{Ablation study on different loss functions for DFD.}
    \label{tab:loss_ablation}
    \begin{tabular}{lcccc}
        \toprule
        \textbf{Method} & MGD \cite{yang2022masked} & Perceptual \cite{johnson2016perceptual} & L1 & \textbf{Ours (MSE)} \\
        \midrule
        \textbf{PSNR (dB)} & 20.04 & 20.34 & 21.99 & \textbf{22.36} \\
        \bottomrule
    \end{tabular}
\end{table}

\section{Limitation and Future Scope}
\label{limit}
Our proposed QDR framework achieves performance comparable to full-precision models at 8-bit (INT8) quantization, effectively preserving high-frequency textures and structural fidelity. However, we observe limitations under ultra-low-bit regimes; specifically, at 4-bit and 2-bit precision, the restored images begin to illustrate noticeable quantization artifacts and texture degradation. Additionally, our current DFD strategy is primarily tailored for CNN-based encoder-decoder architectures, and extending this approach to Vision Transformers (ViTs) requires further investigation. Finally, as our method is currently focused on single-task image restoration, extending it to video or all-in-one restoration settings, evaluating it across diverse hardware platforms, and applying it to a broader range of downstream vision tasks represent promising directions for future work.

\end{document}